\documentclass[lettersize,journal]{IEEEtran}
\usepackage{cite}
\usepackage{amsmath,amssymb,amsfonts}
\usepackage[linesnumbered,ruled]{algorithm2e}
\usepackage{graphicx}
\usepackage{textcomp}
\usepackage{booktabs}
\usepackage{xcolor,colortbl}
\usepackage{amsthm}
\def\BibTeX{{\rm B\kern-.05em{\sc i\kern-.025em b}\kern-.08em
    T\kern-.1667em\lower.7ex\hbox{E}\kern-.125emX}}
\begin{document}
\title{Label Filling via Mixed Supervision for Medical Image Segmentation from Noisy Annotations}
\author{Ming Li, Wei Shen, Qingli Li, Yan Wang
\thanks{Manuscript submitted on July 2023. This work was supported in part by National Natural Science Foundation of China under Grant 62101191 and Grant 61975056, in part by Natural Science Foundation of Shanghai under Grant 21ZR1420800, and in part by the Science and Technology Commission of Shanghai Municipality under Grant 20440713100. }
\thanks{M. Li, Q. Li, and Y. Wang are with Shanghai Key Laboratory of Multidimensional Information Processing, East China Normal University, Shanghai, China (e-mail:10182100316@stu.ecnu.edu.cn, qlli@cs.ecnu.edu.cn, ywang@cee.ecnu.edu.cn). }
\thanks{W. Shen is with MoE Key Lab of Artificial Intelligence, AI Institute, Shanghai Jiao Tong University, Shanghai, China (e-mail: wei.shen@sjtu.edu.cn).}
\thanks{Corresponding author: Y. Wang}}

\markboth{Journal of \LaTeX\ Class Files,~Vol.~14, No.~8, June~2023}%
{Shell \MakeLowercase{\textit{et al.}}: A Sample Article Using IEEEtran.cls for IEEE Journals}

\newcommand{\yan}[1]{ \textcolor{red}{(yan: #1)}  }
\maketitle

\begin{abstract}
The success of medical image segmentation usually requires a large number of high-quality labels. But since the labeling process is usually affected by the raters' varying skill levels and characteristics, the estimated masks provided by different raters usually suffer from high inter-rater variability. In this paper, we propose a simple yet effective \underline{L}abel \underline{F}illing framework, termed as LF-Net, predicting the groundtruth segmentation label given only noisy annotations during training. The fundamental idea of label filling is to supervise the segmentation model by a subset of pixels with trustworthy labels, meanwhile filling labels of other pixels by mixed supervision. More concretely, we propose a qualified majority voting strategy, \emph{i.e.}, a threshold voting scheme is designed to model agreement among raters and the majority-voted labels of the selected subset of pixels are regarded as supervision. To fill labels of other pixels, two types of mixed auxiliary supervision are proposed: a soft label learned from intrinsic structures of noisy annotations, and raters' characteristics labels which propagate individual rater's characteristics information. LF-Net has two main advantages. 1) Training with trustworthy pixels incorporates training with confident supervision, guiding the direction of groundtruth label learning. 2) Two types of mixed supervision prevent over-fitting issues when the network is supervised by a subset of pixels, and guarantee high fidelity with the true label. Results on five datasets of diverse imaging modalities show that our LF-Net boosts segmentation accuracy in all datasets compared with state-of-the-art methods, with even a 7\% improvement in DSC for MS lesion segmentation.
\end{abstract}

\begin{IEEEkeywords}
Medical image segmentation, noisy annotations, synthetic and real-world data
\end{IEEEkeywords}

\section{Introduction}
\label{sec:introduction}
\begin{figure}[t]
\begin{center}
    \includegraphics[width=0.8\linewidth]{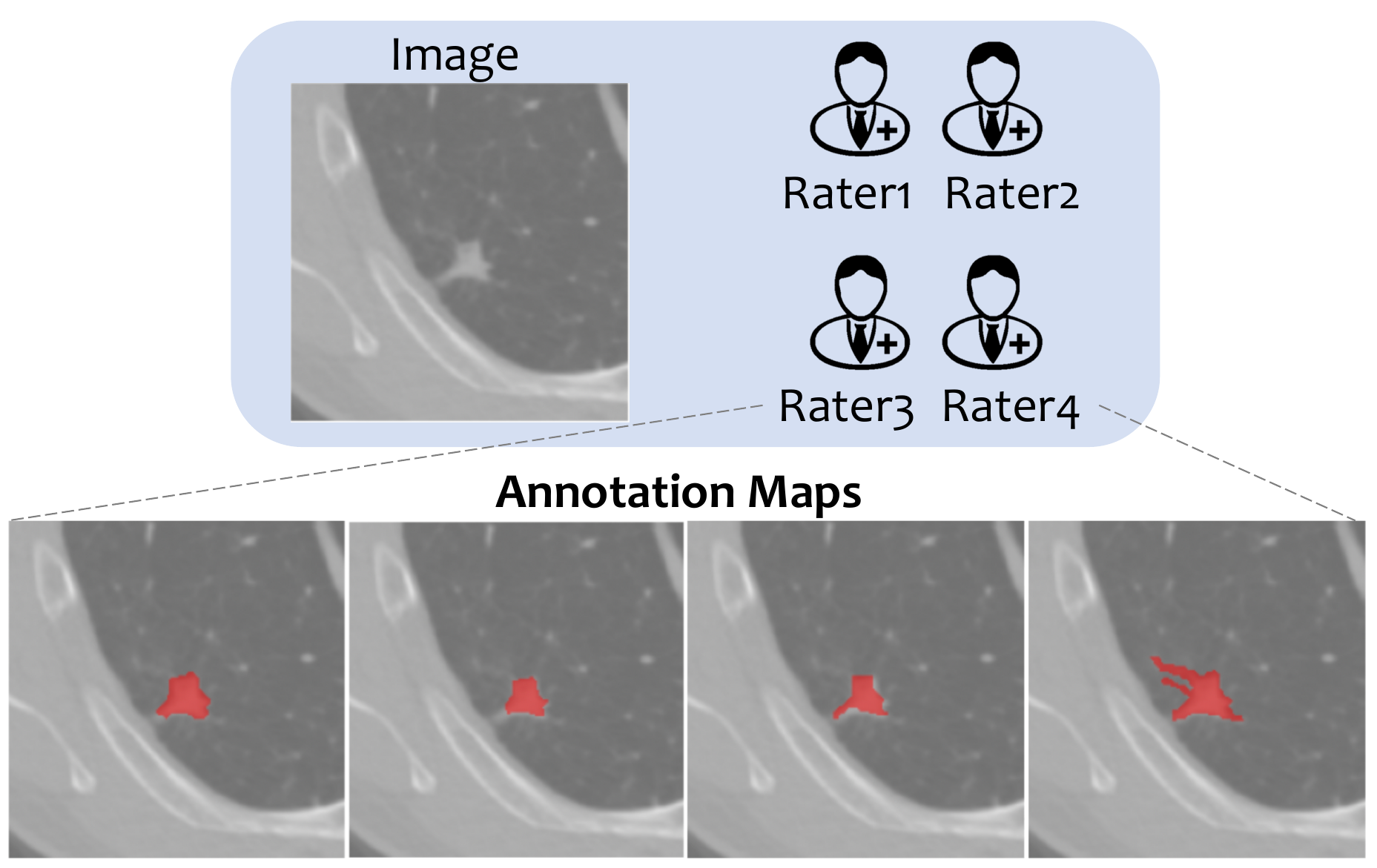}
\end{center}
\caption{A typical label acquisition process. Images and annotation maps are from LIDC-IDRI dataset \cite{SG2011the}.} 
\label{fig:example}
\vspace{-0.5em}
\end{figure}

Machine learning and image segmentation have achieved much success in last decades\cite{sun2021gaussian,kang2017t,rougier2011robust,deng2015single, hu2022cross}. These successes are built on large-scale labeled datasets. However, labeling process is tedious and expensive, and some labeled data may be noisy. When training with small scale dataset, some methods try to design better models for data deficiency\cite{li2023erdunet,shi2023self}.  In this paper, we focus on learning with noisy segmentation annotations.
the lack of agreement amongst human experts and the unsatisfactory quality of labels bring huge challenges for supervised machine learning tasks \cite{Li2021learning,Wang2021knowledge,Wang2020bi,Ding2020semantic,chen2021multi,li2023erdunet,shi2023self}. This challenge is more conspicuous for medical image segmentation, since the manual contouring process is tedious and boundaries of the targets are so uncertain that they always confuse human experts \cite{Wang2021learning,Zhang2020collaborative}.   As mentioned in \cite{Zhang2020disentangling}, in a typical label acquisition process (see an example in Fig.~\ref{fig:example}), a group of experts provide their estimated segmentation masks under the influence of their varying skill-levels, area of expertise, and rater specific biases, such as structural errors \cite{Warfield2004simultaneous}. The creation of well-annotated large scale medical datasets is required for building robust machine learning algorithms but extremely labor intensive \cite{Fagundo2018deep}. Considering the limited data access \cite{Yang2020bi} and the lack of well-annotated labeling in most research groups and industry \cite{Willemink2020preparing}, studying noisy annotations from multiple raters is necessary for the success of machine learning in radiology and pathology.

In this paper, following \cite{Zhang2020disentangling}, we investigate supervised medical image segmentation that predicts the true segmentation labels, given only noisy annotations. Majority voting (also called relative majority voting, meaning winner-takes-all), STAPLE \cite{Warfield2004simultaneous} and other label fusion methods \cite{Menze2015the,Asman2011robust} are the most basic strategies to curate segmentation annotations when embracing annotations from multiple raters. But these methods fail to model the relationships among reliability, noisy annotation patterns of raters and the underlying true labels. When the level of annotation errors is high, they may result in inferior segmentation performances. In addition, the model may be prone to be over-confident \cite{Ji2021learning} if trained with the typical groundtruth label via \emph{e.g.}, majority vote \cite{Ji2021learning}. Recently, human error disentangling frameworks \cite{Zhang2020disentangling,Tanno2019learning} assume there is a single, unknown, true segmentation map and a group of raters offer their individual estimations \emph{w.r.t.} own biases.  

We follow the aforementioned research direction and hold the assumption that in the context of segmentation problems, there exists one underlying true segmentation mask that is needed to explore given noisy labels.  Methods in \cite{Zhang2020disentangling,Tanno2019learning} simply treat the underlying true label as latent values, and map the input image to the noisy annotations. Human biases are delicately modeled, but they suffer from two drawbacks: (1) The characteristics of individual raters are modeled by estimating pixel-wise \underline{C}onfusion \underline{M}atrices (CM), which inevitably ignores the neighboring context of pixels. (2) They assume estimated and true CMs should be diagonally dominant, so that the estimated CMs uniquely recovers the true CMs. But in real scenarios, not all pixels' CMs are diagonally dominant, since some arbitrary pixels (\emph{e.g.}, on the boundary of the target) may be difficult to label. Besides, the learning process by SGD cannot guarantee that the diagonally dominant assumption is always held. This may result in inferior performances.

\begin{figure}[t]
\begin{center}
    \includegraphics[width=1\linewidth]{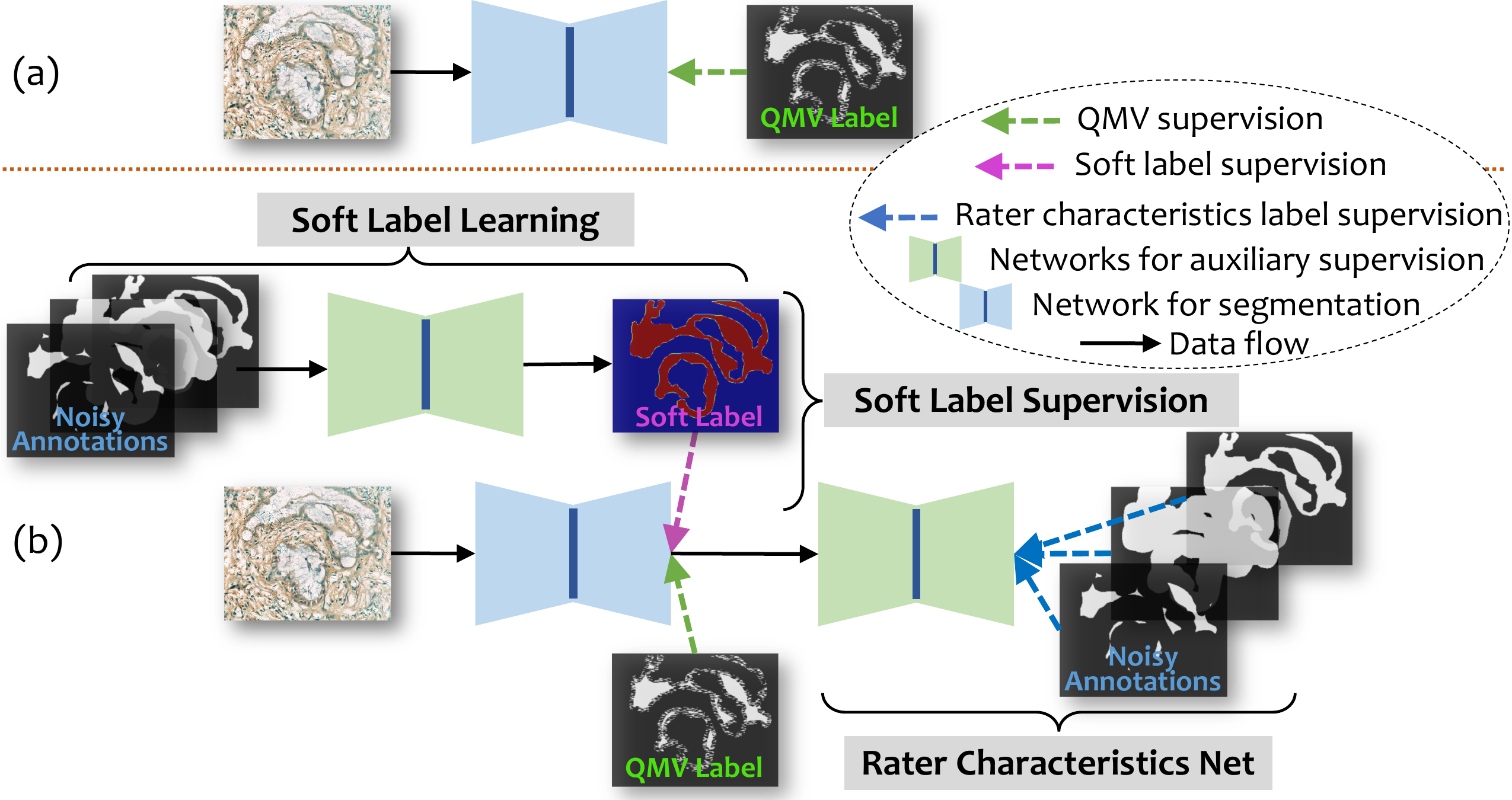}
\end{center}
\caption{A glimpse of designed architectures. (a) A segmentation network with the proposed QMV strategy. (b) A segmentation network with mixed supervision. QMV means qualified majority voting.} 
\label{fig:illustration}
\vspace{-0.5em}
\end{figure}

We propose a simple, clean yet effective model, called \underline{L}abel \underline{F}illing Net (LF-Net). Our motivation comes from a simple observation that the probability of making mistakes if we annotate one pixel with a specific class that nearly all raters agree with is very small. 
Thus, we can trust the label of this pixel if more than a certain number of raters agree as its groundtruth label, which is also termed as qualified majority voting. 
Then, how to fill true labels for other pixels is an open issue. To address this problem, we propose mixed supervision by learning raters' agreement and individual rater's characteristics. More specifically, we first propose a threshold voting scheme to model agreement among raters and regard the labels of the most trustworthy pixels as the groundtruth, as shown in Fig.~\ref{fig:illustration} (a). This supervision is also known as qualified majority voting label. But, if a network is only supervised by the selected pixels in training, it will inevitably face the problem of over-fitting, leading to sub-optimal solutions. Besides, the information from other pixels will be lost if only focusing on a subset of pixels. We design two types of auxiliary supervision to alleviate these issues, as shown in Fig.~\ref{fig:illustration} (b):

\begin{itemize}
    \item[(1)] A soft label learned from intrinsic structures of noisy annotations. Given noisy annotations, a soft label learning network is designed to construct a fused soft label, and it is supervised by the designed qualified majority voting. Though supervised by a subset of pixels, the soft label can be filled by learning the intrinsic structures of noisy annotations via deep networks. The consensus captured by the soft label can be considered as a denoised annotation. We further propose to supervise the segmentation network by the rich representation capability of the soft label. Experimental results show that compared with other label fusion methods, our learning-based soft label works more favorably for guiding the subsequent true label learning, leading to significant performance boost.
    \item[(2)] Raters' characteristics labels (\emph{i.e.}, noisy annotations) which propagate individual rater's characteristics information. For datasets that have consistent rater IDs across images, a rater characteristics net module is added into the segmentation network, which models the characteristics of individual raters based upon the learned true label. Thereon, the predicted true label in the segmentation network can be filled through incorporating information propagated by raters' characteristics.   
\end{itemize}

We verify LF-Net on five datasets, which are from varied image modalities, including handwritten digits, color fundus images, \underline{C}omputed \underline{T}omography (CT) and \underline{H}yper\underline{S}pectral \underline{I}mages (HSI). Strictly following \cite{Zhang2020disentangling}, multi-rater annotations are simulated by performing morphometric operations for three datasets: MNIST, ISBI2015 MS lesion segmentation challenge dataset \cite{Jesson2015hierarchical} and multi-dimensional choledoch dataset for cholangiocarcinoma diagnosis \cite{Zhang2019a}. For the other two datasets: LIDC-IDRI \cite{SG2011the} and RIGA \cite{Almazroa2017agreement}, we adopt their real experts' labels as multi-rater annotations. Since LIDC-IDRI does not have consistent rater IDs, we also analyze the results on LF-Net without rater characteristics labels. Without bells and whistles, our LF-Net consistently exceeds all state-of-the-art methods by a large margin, with even 7\% improvement in terms of Dice-S{\o}rensen coefficient for MS lesion segmentation. The ablation study further shows the effectiveness of each proposed module in LF-Net.

\section{Related Work}
\subsection{Multi-rater Modeling}
\subsubsection{Multi-rater with Ambiguous Groundtruth}
We will introduce two research directions \emph{w.r.t.} whether there exists a single groundtruth label in multi-rater modeling. The first one is multi-rater with ambiguous groundtruth. In this case, one may need to train a model that can reflect the inter-rater variability. \cite{Kohl2018a,Kohl2019a} proposes a generative model based on U-Net \cite{Ronneberger2015unet} and a conditional variational autoencoder \cite{Sohn2015learning} to learn a distribution over segmentation. In these works, a {G}eneralised {E}nergy {D}istance is used to evaluate the segmentation synthesized by their models. \cite{Jensen2019improving} designs a label sampling strategy, which samples labels randomly from the multi-rater labeling pool during training. But, corrupted annotation from each annotator can degrade the model's final performance. Recently, MR-Net, a two-stage method which models multi-rater's (dis-)agreement is proposed \cite{Ji2021learning}. The goal of \cite{Ji2021learning} is to be capable of giving calibrated predictions. The soft dice coefficient and the soft Intersection Over Union metrics through multiple thresholds are used. 

\subsubsection{Multi-rater with Single True Groundtruth}
Another line of work assumes there is only one objective groundtruth, and all annotations provided could be noisy. The goal is to leverage information from raters' annotations and prevent noise to degrade the model. \cite{Guan2018who} first learns to predict the annotation of each rater and then learns a corresponding weight for final prediction. \cite{Yu2020difficulty} presents a multi-branch structure to leverage multi-rater consensus information for glaucoma classification. Recently, \cite{Zhang2020disentangling} designs two coupled CNNs to learn true predictions and the confusion matrix for each pixel, whose idea is similar with \cite{Tanno2019learning} for classification. They assume that annotations over different pixels are independent given the input image, which are not practical in real scenarios. Our work is related with \cite{Zhang2020disentangling,Tanno2019learning,chang2003cbsa,chen2021multi}, but unlike these methods, which simply treat the groundtruth as latent variables. We propose a label filling method, which chooses the most trustworthy pixels to guide the training direction of the segmentation model. It ensures that our segmentation, especially for the selected pixels is in line with the groundtruth. Besides, our method incorporates rich agreement information from noisy annotations, which alleviates the over-fitting issue. Thus, the rest of the pixels can be filled with correct labels.

\subsection{Agreement Modeling}
Agreement modeling is an important technique in machine learning. For example, \cite{zhu2021generalizable} proposes to measure the gradient agreement among N NR-IQA tasks with synthetic distortions, and \cite{ge2022self} 
proposes to employ K autoencoders to learn K subspaces and summarise the level of agreement between K observers. For segmentation, traditional methods of label aggregation such as STAPLE \cite{Warfield2004simultaneous} and its extensions \cite{Asman2011robust,Asman2012formulating,Cardoso2013steps,Asman2013non,Joskowicz2018automatic} always resort to treat the groundtruth as latent variables. However, some of these methods assume the noisy annotations are provided during testing, which is different from ours. Very recently, \cite{Wang2021agreementlearning} uses a two-stream method which learns information from all annotations and models raters' agreement as regularization. The agreement learning strategy proposed in \cite{Wang2021agreementlearning} assumes the groundtruth is ambiguous to define, and the objective is to improve the overall agreement between the model and the annotators. Thus, different from ours, they evaluate the model without groundtruth.

\section{Methodology}
\subsection{Problem Formulation}
\label{sec:pf}
In the context of segmentation problems \cite{Wang2020deep,Zhao20213d,Liu2020msnet}, one aims at finding one underlying true segmentation mask that is needed to explore given noisy annotations \cite{Zhang2020disentangling,Tanno2019learning}. Following \cite{Zhang2020disentangling}, we define a set of $M$ images $\{\mathbf{X}_m\in\mathbb{R}^{W\times H\times C}\}_{m=1}^M$, where $W$, $H$, and $C$ denote the width, height, and channel of the image, respectively. For every image $\mathbf{X}_m$, a set of noisy segmentation labels $\{\widetilde{\mathbf{Y}}_m^r\in\mathbb{Y}^{W\times H}\}_{r\in \mathcal{E}(\mathbf{X}_m)}$ are given, where $\widetilde{\mathbf{Y}}_m^r$ denotes the segmentation labels from $r$th rater $r\in\{1,2,...,R\}$, $\mathcal{E}(\mathbf{X}_m)$ denotes the set of raters who labeled image $\mathbf{X}_m$, and $\mathbb{Y}=[1, 2, ..., L]$ denotes the set of classes. We assume that no groundtruth labels $\{\mathbf{Y}_m\in\mathbb{Y}^{W\times H}\}_{m=1}^M$ are available and $|\mathcal{E}(\mathbf{X}_m)|=R$. We will drop subscript $m$ in $\mathbf{X}_m$, $\tilde{\mathbf{Y}^r_m}$ and $\mathbf{Y}_m$ for simplicity. Our goal is to predict the underlying true label $\mathbf{Y}_t$ for a testing image $\mathbf{X}_t$, by leveraging useful information in noisy segmentation labels to train our model and meanwhile, preventing noisy labels to degrade our model.

\subsection{Prerequisite and Motivation}
\label{sec:pre}

Although the segmentation labels annotated by raters may be noisy, it is infrequent that most raters make mistakes simultaneously. Assume that we have $R$ raters to annotate one image. Without loss of generality, for any pixel $s$ in the image, the true (groundtruth) label of it is defined as $y_{s}$, and $y_{s}\in\{1,2,...,L\}$. Assume that for the $r$th rater (expert) $e_r$, where $r\in\{1,2,...,R\}$, the probability of labeling pixel $s$ correctly is $p_r=p(\tilde{y}_s^r=l|y_s=l)$.
where $l\in\{1,2, ..., L\}$, and $\tilde{y}_s^r$ is the annotation made by rater $e_r$. We assume that the process of raters' annotations is independent, and the probability of pixel $s$ belonging to class $l$ is $\frac{1}{L}$ if we do not have any prior expertness information, \emph{i.e.},
\begin{equation}
    p(\tilde{y}_s^{1}, \tilde{y}_s^{2}, ..., \tilde{y}_s^{R}|y_s) = p(\tilde{y}_s^{1}|y_s)p(\tilde{y}_s^{2}|y_s)...p(\tilde{y}_s^{R}|y_s),
\end{equation}
where $\tilde{y}_s^{1}$, ..., $\tilde{y}_s^{R}\in\{1,2,...,L\}$ and $p(y_s=l)=1/L$. If $T$ raters label $s$ to a specific class $\tilde{l}$ and the remaining $R-T$ raters contend $s$ should be labeled with other classes instead of $\tilde{l}$, \emph{i.e.},
\begin{equation}
    \tilde{y}_s^{x_1}=...=\tilde{y}_s^{x_T}=\tilde{l}, \tilde{y}_s^{x_{T+1}},...,\tilde{y}_s^{x_R}\in\{1,...,L\}/\{\tilde{l}\},
\end{equation}
where $x_{1},...x_{R}$ are permutations of $1,...,R$. In other words, we focus on the pixels that most raters' annotations are the same, \emph{i.e.}, $T=R-C$, where $C\in\mathbb{N}$, which is a small value, and $\mathbb{N}$ indicates the set of non-negative integers. Note that we do not add any assumption about the order and annotation about one specific rater. We now consider the following Lemma: 

\noindent\textbf{Lemma.} $\forall r=1,2,...,R$, $p(\tilde{y}_s^{r}=l|y_s\neq l)\leq 1-p_r$, where $l\in\{1,2,...,L\}$.

\noindent With this Lemma, we have the following theorem hold (\emph{proof} is provided in \emph{Appendix}):

\noindent \textbf{Theorem.} Assume for a pixel $s$, $p_{max}=\max\{p_1, p_2, ...,p_R\}<1$, $p_{min} = \min\{p_1,p_2,...,p_R\}>1/2$, and $\forall r=1,2,...,R$, $p(\tilde{y}_s^{r}\neq\tilde{l}|y_s=\tilde{l})>0$, where $p_r$ indicates the probability of labeling pixel $s$ correctly. $T$ raters label $s$ to class $\tilde{l}$. $T = R-C$, where $C$ is a constant, and $C\in\mathbb{N}$.  We define the probability that $y_s=\tilde{l}$ given R raters' annotations is ${P}_{gt}(R) = p(y_s=\tilde{l}|\tilde{y}_s^{x_1}=...=\tilde{y}_s^{x_T}=\tilde{l}, \tilde{y}_s^{x_{T+1}}, ..., \tilde{y}_s^{x_R}\in\{1,...,L\}/\{\tilde{l}\}, T=R-C)$. Then $\lim_{R\rightarrow +\infty}{P}_{gt}(R)=1$.

\begin{figure*}[t]
\begin{center}
    \includegraphics[width=0.9\linewidth]{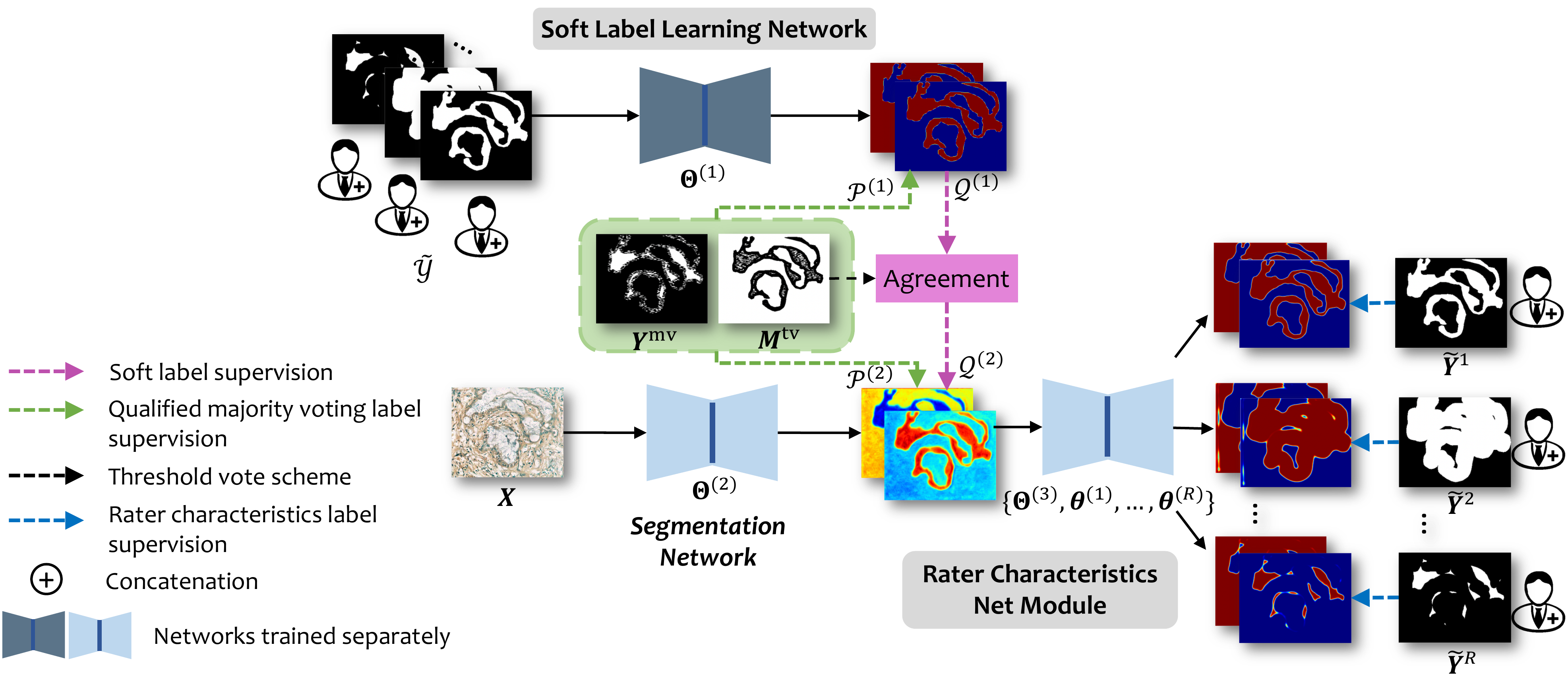}
\end{center}
\vspace{-0.9em}
\caption{The training stage of LF-Net. To guide the training process in line with the groundtruth, we propose a qualified majority voting strategy (a threshold voting scheme selects pixels whose majority-voted labels are trustworthy). First, a soft label learning network is trained. Then, to assist the learning of the segmentation model and fill correct labels of other pixels, two types of auxiliary supervision are designed: (1) the soft label obtained from soft label learning network is armed with information from intrinsic structures of noisy annotations, and (2) rater characteristics labels (\emph{i.e.}, noisy annotations) disentangling human errors from filled label by rater characteristics net module, so as to help the label filling when back-propagating.} 
\label{fig:framework}
\vspace{-1em}
\end{figure*}

For a two-class case, \emph{i.e.}, $l\in\{0,1\}$, when $C=1$, $p_{max}=0.95$,  and $p_{min}=0.75$, one can easily compute ${P}_{gt}(7)\geq 0.975$, ${P}_{gt}(8)\geq0.991$, and ${P}_{gt}(9)\geq0.997$ (detailed derivation is provided in \emph{Appendix}). Thereon, the annotation that provided through this qualified majority voting is termed as \emph{trustworthy} annotation throughout the paper. 

\emph{The above prerequisite motivates us to propose a threshold voting scheme.} We propose a mask $\mathbf{M}^\text{tv}$ (see Eq.~\ref{eq:tv}), which selects a subset of pixels whose majority-voted labels are trustworthy as the supervision, to guide the true label learning of the network. Theoretically, let $\hat{h}$ be the optimal hypothesis from the input image to the unknown groundtruth. To approximate $\hat{h}$, a segmentation model determines a hypothesis space $\mathbb{H}$ of hypotheses. It should be emphasized that with the help of the threshold voting scheme, we constrain the complexity of $\mathbb{H}$, which results in a much smaller hypothesis space $\tilde{\mathbb{H}}$ \cite{Wang2020generalizing}. With the prerequisite in Sec.~\ref{sec:pre}, we know that $\hat{h}\in\tilde{\mathbb{H}}$. 

But, if one network is only supervised by the selected pixels, it may face the problem of over-fitting (see Fig.~\ref{fig:illustration}) (a), since the labeled pixels are a small amount of easy samples. Thus, it may fail to fill correct labels for other pixels. To mitigate this issue, we propose the following framework for medical image segmentation.

\subsection{Overall Framework}
\label{sec:framwork}
In this work, we propose LF-Net. As shown in Fig.~\ref{fig:framework}, the proposed LF-Net first constructs a fused soft label by modeling raters' agreement (see Sec.~\ref{sec:lal}), 
which provides the segmentation network with more informative labels (see Sec.~\ref{sec:ad})). This makes the training easier and prevent over-fitting. The segmentation network, which acquires agreement knowledge from the well-trained soft label learning network, is also armed with extra knowledge from individual rater's characteristics. The rater characteristics module attached to the segmentation network describes the noisy labelling process based on the true label (see Sec.~\ref{sec:rcn}). Thus, the learned raters' characteristics can be back-propagated, and the true label learning ability can be enhanced. 

\subsection{Soft Label Learning Network}
\label{sec:lal}

The main purpose of soft label learning network is to learn a fused soft label from noisy annotations $\{\tilde{\mathbf{Y}}^r\}_{r\in \mathcal{E}(\mathbf{X})}$, 
so that the information learned from noisy annotations can be used to supervise the segmentation network. 
Regarding other label fusion choices, majority voting, STAPLE or averaging cannot take noisy patterns such as contextual information into consideration. We shall see in the experiments that other label fusion methods are unable to improve segmentation accuracy over our learning-based label fusion method.

We concatenate $R$ raters' annotations $\{\tilde{\mathbf{Y}}^r\}_{r\in \mathcal{E}(\mathbf{X})}$ into a tensor $\tilde{\mathcal{Y}}\in \mathbb{Y}^{W\times H\times R}$, where the vector on spatial position $(i,j)$ is defined as $\tilde{\mathbf{y}}_{i,j}$. 
It is fed to the network to model agreement via exploiting contextual spatial information. As shown in Fig. \ref{fig:framework}, our soft label learning network is supervised by majority voted label $\mathbf{Y}^{\text{mv}}=\Omega(\tilde{\mathcal{Y}})\in\mathbb{Y}^{W\times H}$ for each spatial location, where $\Omega$ indicates the majority vote function to choose a simple majority label for each pixel. As shown in Sec.~\ref{sec:pre}, the annotation that provided by qualified majority voting is trustworthy, we define a new threshold voting scheme which can obtain a $0-1$ mask $\mathbf{M}^\text{tv}\in \mathbb{Z}^{W\times H}$, where $\mathbb{Z}=\{0,1\}$, and the value on position $(i,j)$ is defined as $m_{i,j}^\text{tv}$:
\begin{equation}
\label{eq:tv}
    m_{i,j}^\text{tv}=\Psi(\tilde{\mathbf{y}}_{i,j})=
\begin{cases}
1 &  \max\{N_{i,j}^1, ... N_{i,j}^L\}\geq \beta\\
0 &  \text{otherwise},\\
\end{cases}
\end{equation}
where $N_{i,j}^l$ denotes the number of raters annotating class $l$ for location $(i,j)$. In our experiments, we set the threshold $\beta=R-1$.
The loss function $\mathcal{L}_\text{sll}$ for training our soft label learning network is defined as:
\begin{equation}
\label{eq:sll}
    \mathcal{L}_\text{sll}(\mathbf{\Theta}^{(1)})=-\sum_{i,j}\sum_{l=1}^L\left(m_{i,j}^\text{tv}\cdot\mathbf{1}(y_{i,j}^\text{mv}=l)\log {p}_{i,j,l}^{(1)}\right),
\end{equation}
where $\mathbf{1}(\cdot)$ is an indicator function, $\mathbf{\Theta}^{(1)}$ is the parameter of the soft label learning network, $y_{i,j}^\text{mv}$ is the value on position $(i,j)$ of $\mathbf{Y}^\text{mv}$, and 
the probability map of the soft label learning network is defined as a tensor ${\mathcal{P}}^{(1)}\in \mathbb{R}^{W\times H\times L}$, where the value on location $(i,j,l)$ is ${p}^{(1)}_{i,j,l}$.
After the network converges, the optimal parameter is obtained by
\begin{equation}
(\hat{\mathbf{\Theta}}^{(1)})=\arg\min_{{\mathbf{\Theta}}^{(1)}}{\mathcal{L}_{\text{sll}}({\mathbf{\Theta}}^{(1)})}.
\end{equation}
The soft label is then computed by $\hat{\mathcal{P}}^{(1)}=f(\tilde{\mathcal{Y}};\hat{\mathbf{\Theta}}^{(1)})$, where $f(\cdot; \mathbf{\Theta})$ is a segmentation network parameterized by $\mathbf{\Theta}$. 

\subsection{The Segmentation Network with Mixed Supervision}

Next, we introduce our proposed segmentation network with mixed supervision. As shown in Fig.~\ref{fig:illustration} (a), a straightforward strategy to train a segmentation network is to use qualified majority voting as supervision. Mathematically, let $\mathbf{\Theta}^{(2)}$ parameterizes a vanilla segmentation network. The segmentation network outputs a probability map $\mathcal{P}^{(2)}=f(\mathbf{X};\mathbf{\Theta}^{(2)})$, and the value on location $(i,j,l)$ is $p_{i,j,l}^{(2)}$. The proposed loss function for supervising $\mathcal{P}^{(2)}$ in the vanilla segmentation network is:
\begin{equation}
    \mathcal{L}_\text{ce}(\mathbf{\Theta}^{(2)})=-\sum_{i,j}\sum_{l=1}^L\left(m_{i,j}^\text{tv}\cdot\mathbf{1}(y_{i,j}^\text{mv}=l)\log {p}_{i,j,l}^{(2)}\right).
\end{equation}

To make training easier and prevent the over-fitting issue of the segmentation network, as discussed in Sec.~\ref{sec:framwork}, we propose two types of auxiliary supervision: (1) soft label supervision and (2) rater characteristics supervision by rater characteristics net module.

\subsubsection{Soft Label Supervision}
\label{sec:ad}
The soft label learning is supervised by only a subset of pixels which receive consensus among most annotators. This supervised annotation can be considered as a denoised annotation, and we train deep networks to learn such a denoised trend given noisy annotations. Thus, supervising the segmentation network by the learned soft label is helpful for the label filling task. 

The soft label learning model outputs a probability map $\hat{\mathcal{P}}^{(1)}=f(\mathbf{X};\hat{\mathbf{\Theta}}^{(1)})$. We define $\hat{p}_{i,j,l}^{(1)}=\sigma\left(\hat{a}_{i,j,l}^{(1)}\right)=\frac{\exp\left(\hat{a}_{i,j,l}^{(1)}\right)}{\sum_{t=1}^L\exp\left(\hat{a}_{i,j,t}^{(1)}\right)}$, where $\hat{a}_{i,j,l}^{(1)}$ is the activation value of the $(i,j)$th pixel on the $l$th channel dimension. 

To train the network by soft label supervision, we first adopt softmax with temperature $\tau$ on the activation value $\hat{a}_{i,j,l}^{(1)}$ to soften the vanilla softmax distribution,
which is given:
\begin{equation}
\label{eq:kl}
    \hat{q}_{i,j,l}^{(1)}=\sigma_\text{soft}(\hat{a}_{i,j,l}^{(1)})=\frac{\exp\left(\hat{a}_{i,j,l}^{(1)}/\tau\right)}{\sum_{t=1}^L\exp\left(\hat{a}_{i,j,t}^{(1)}/\tau\right)},
\end{equation}
and $\hat{q}_{i,j,l}^{(1)}$ is the value on position $(i,j,l)$ of tensor $\hat{\mathcal{Q}}^{(1)}\in\mathbb{R}^{W\times H\times L}$. Subsequently, we propose a constraint KL-divergence loss as the \underline{S}oft \underline{L}abel (SL) supervision loss $\mathcal{L}_\text{sl}$:
\begin{equation}
\label{eq:sls}
    \mathcal{L}_\text{sl}(\mathbf{\Theta}^{(2)})=-\tau^2\sum_{i,j}\sum_{l=1}^L\left(m_{i,j}^\text{tv}\cdot \hat{q}_{i,j,l}^{(1)}\log q_{i,j,l}^{(2)}\right),
\end{equation}
where $q_{i,j,l}^{(2)}$ denotes the softened probability from the segmentation network with temperature $\tau$, and $q_{i,j,l}^{(2)}$ denotes the value on location $(i,j,l)$ of tensor $\mathcal{Q}^{(2)}\in\mathbb{R}^{W\times H\times L}$. With a threshold voting scheme, the soft label supervision loss provides the segmentation network with representation capability to model the underlying true label from agreement representation.

\subsubsection{{R}ater Characteristics Label Supervision}
\label{sec:rcn}

The rater characteristics net module characterizes the process that each rater produces a noisy approximation to the underlying true label. Different from \cite{Zhang2020disentangling,Tanno2019learning}, which assume that annotations over different pixels are independent, we consider the spatial contextual information of nearby pixels, and model the individual characteristics through deep networks. Our rater characteristics net module, parametered by $\{\mathbf{\Theta}^{(3)}, \boldsymbol{\theta}^{(1)}, ..., \boldsymbol{\theta}^{(R)}\}$, generates $R$ segmentation probability maps corresponding to $R$ raters respectively, where $\mathbf{\Theta}^{(3)}$ is the parameters of the network backbone. RCN module has $R$ head branches, each of which is paramterized by $\boldsymbol{\theta}^{(r)}$, targeting on the noisy labels provided by the $r$th rater. Thus, these branches perform segmentation with \underline{R}ater \underline{C}haracteristics \underline{L}abel (RCL) supervision loss function:
\begin{equation}
    \mathcal{L}_\text{rcl}(\mathbf{\Theta}^{(3)}, \boldsymbol{\theta}^{(1)}, ..., \boldsymbol{\theta}^{(R)})=-\sum_{r=1}^R\sum_{i,j}\sum_{l=1}^L\left(\mathbf{1}(\tilde{y}_{i,j}^r=l)\log p_{i,j,l}^{r}\right),
\end{equation}
where $p_{i,j,l}^r$ is the value on location $(i,j,l)$ of $r$th segmentation probability map. 

Finally, with the soft label supervision and the rater characteristics label supervision, the loss function for training the segmentation network is
\begin{equation}
\label{eq:seg_loss}
\mathcal{L}_\text{seg}=\mathcal{L}_\text{sl}(\mathbf{\Theta}^{(2)})+\mathcal{L}_\text{ce}(\mathbf{\Theta}^{(2)})+\lambda\mathcal{L}_\text{rcl}(\mathbf{\Theta}^{(3)}, \boldsymbol{\theta}^{(1)}, ..., \boldsymbol{\theta}^{(R)}),
\end{equation}
where $\lambda$ is a trade-off parameter. The overall training procedure is summarized in Algorithm~\ref{Alg:train}, where we assume the batchsize is $1$ for simplicity.

\begin{algorithm}[t!]
\small
\SetKwInOut{Input}{Input}
\SetKwInOut{Output}{Output}
\SetKwInOut{Return}{Return}
\Input{
Training set $\mathcal{D}=\left\{(\mathbf{X}_m,\{\widetilde{\mathbf{Y}}_m^r\}_{r\in \mathcal{E}(\mathbf{X}_m)})\right\}_{m=1}^M$;\\
Max number of iterations $T_1$ and $T_2$;\\
}
\Output{
Parameters $\hat{\mathbf{\Theta}}^{(2)}$;
}
${t}\leftarrow{0}$;\\
Randomly initialize $\mathbf{\Theta}^{(1)}$, $\mathbf{\Theta}^{(2)}$ $\mathbf{\Theta}^{(3)}$ and $\{\boldsymbol{\theta}^{(1)},...,\boldsymbol{\theta}^{(R)}\}$;\\
\Repeat{${t}={T_1}$}
{${t}\leftarrow{t+1}$;\\
Randomly select a data sample $(\mathbf{X},\{\widetilde{\mathbf{Y}}^r\}_{r\in \mathcal{E}(\mathbf{X})})$ from $\mathcal{D}$;\\
Compute $\mathcal{L}_\text{sll}(\mathbf{\Theta}^{(1)})$ on the data sample by Eq.~\ref{eq:sll};\\
Update $\mathbf{\Theta}^{(1)}$ by Gradient Descent;\\
}
$\hat{\mathbf{\Theta}}^{(1)}\leftarrow\mathbf{\Theta}^{(1)}$;\\
${t}\leftarrow{0}$;\\
\Repeat{${t}={T_2}$}
{
$t\leftarrow t+1$;\\
Randomly select a data sample $(\mathbf{X},\{\widetilde{\mathbf{Y}}^r\}_{r\in \mathcal{E}(\mathbf{X})})$ from $\mathcal{D}$;\\
Obtain $\hat{\mathcal{Q}}^{(1)}$ by Eq.~\ref{eq:kl};\\
Compute $\mathcal{L}_\text{seg}$ by Eq.~\ref{eq:seg_loss};\\
Update $\mathbf{\Theta}^{(2)}$, $\mathbf{\Theta}^{(3)}$, and $\{\boldsymbol{\theta}^{(1)},...,\boldsymbol{\theta}^{(R)}\}$ by Gradient Descend;\\
}
\Return{
$\hat{\mathbf{\Theta}}^{(2)}\leftarrow \mathbf{\Theta}^{(2)}$
}
\caption{
Training process for LF-Net, with batchsize $=1$
}
\label{Alg:train}
\end{algorithm}

During testing, an image $\mathbf{Z}$ is fed into the segmentation network (\emph{i.e.}, parameterized by $\mathbf{\Theta}^{(2)}$) to obtain the final segmentation prediction.

\begin{table*}[!tb]
\renewcommand\arraystretch{1}
\footnotesize
\centering
\caption{Performance comparison (in DSC, $\%$ and bAHD) on MNIST, MS Lesion \cite{Jesson2015hierarchical} and Choledoch \cite{Zhang2019a} datasets. Experiments are repeated at least 3 times with different model initializations to compute the mean and standard deviation, as suggested in \cite{Zhang2020disentangling}. ``$\uparrow$" and ``$\downarrow$" indicate the larger and the smaller the better, respectively. QMV-UNet indicates using our proposed qualified majority voting scheme. \textbf{Bold} denotes the best results for each dataset per measurement. \emph{Average} represents the average DSC of all annotations in training dataset, indicating the noisy level of annotations. \emph{Oracle} means our LF-Net is trained with all groundtruth annotations, served as the upper bound.}
\label{tab:comparison}
\begin{tabular}{lcccccc}
\toprule[0.15em]
&  \multicolumn{2}{c}{MNIST} & \multicolumn{2}{c}{MS Lesion \cite{Jesson2015hierarchical}} & \multicolumn{2}{c}{Choledoch \cite{Zhang2019a}}\\
\cmidrule{2-3} \cmidrule(lr){4-5} \cmidrule(lr){6-7}
{Methods} & DSC~$\uparrow$ & bAHD~$\downarrow$ & DSC~$\uparrow$ & bAHD~$\downarrow$ & DSC~$\uparrow$ &  bAHD~$\downarrow$ \\ 
\midrule[0.09em]
\emph{Average} & {47.10} & N/A & {47.22} & N/A & {51.97} & N/A \\
\arrayrulecolor{black!30}\midrule
MV-UNet & 57.62 $\pm$ 0.44 & 1.065 $\pm$ 0.015 & 42.48 $\pm$ 0.53 & 1.91 $\pm$ 0.06 & 36.87 $\pm$ 0.73 & 54.71 $\pm$ 3.58 \\
QMV-UNet & 75.36 $\pm$ 0.48 & 0.429 $\pm$ 0.085 & 46.32 $\pm$ 1.06 & 3.27 $\pm$ 0.29 & 50.07 $\pm$ 0.52 & 35.31 $\pm$ 1.94\\
Mean-UNet & 59.09 $\pm$ 0.54 & 0.658 $\pm$ 0.090 & 44.25 $\pm$ 1.34 & 2.07 $\pm$ 0.40 & 43.07 $\pm$ 0.41 & 43.18 $\pm$ 2.30 \\
STAPLE-UNet & 90.49 $\pm$ 0.45 & 0.066 $\pm$ 0.012 & 60.98 $\pm$ 1.19 & 1.66 $\pm$ 0.26 & 58.14 $\pm$ 0.12 & 37.18 $\pm$ 1.55 \\
LS-UNet \cite{Jensen2019improving} & 66.30 $\pm$ 1.31 & 0.518 $\pm$ 0.067 & 43.40 $\pm$ 1.74 & 4.57 $\pm$ 3.24 & 46.49 $\pm$ 4.36 & 67.16 $\pm$ 7.33 \\
MH-UNet \cite{Guan2018who} & 79.80 $\pm$ 0.53 & 0.310 $\pm$ 0.108 & 50.62 $\pm$ 0.65 & 1.86 $\pm$ 0.39 & 51.71 $\pm$ 0.20 & 38.05 $\pm$ 1.44 \\
CM-UNet \cite{Zhang2020disentangling} & 93.09 $\pm$ 0.19 & 0.033 $\pm$ 0.011 & 72.45 $\pm$ 0.31 & 1.08 $\pm$ 0.14 & 61.43 $\pm$ 0.38 & 32.53 $\pm$ 1.31\\
LF-Net (ours) & \textbf{99.25} $\pm$ \textbf{0.17} & \textbf{0.009} $\pm$ \textbf{0.001} & \textbf{79.08} $\pm$ \textbf{0.24} & \textbf{0.65} $\pm$ \textbf{0.13} & \textbf{65.35} $\pm$ \textbf{0.33} & \textbf{23.22} $\pm$ \textbf{1.84} \\
\arrayrulecolor{black!30}\midrule
\emph{Oracle} & 99.58 $\pm$ 0.18 & 0.005 $\pm$ 0.001 & 80.39 $\pm$ 0.51 & 0.54 $\pm$ 0.13 & 67.02 $\pm$ 0.35 & 21.96 $\pm$ 1.32 \\
\bottomrule[0.15em]
\end{tabular}
\end{table*}

\section{Experimental Results}
\label{sec:exp}

\subsection{Dataset, Split and Pre-processing}
\begin{figure}[t]
\begin{center}
    \includegraphics[width=1\linewidth]{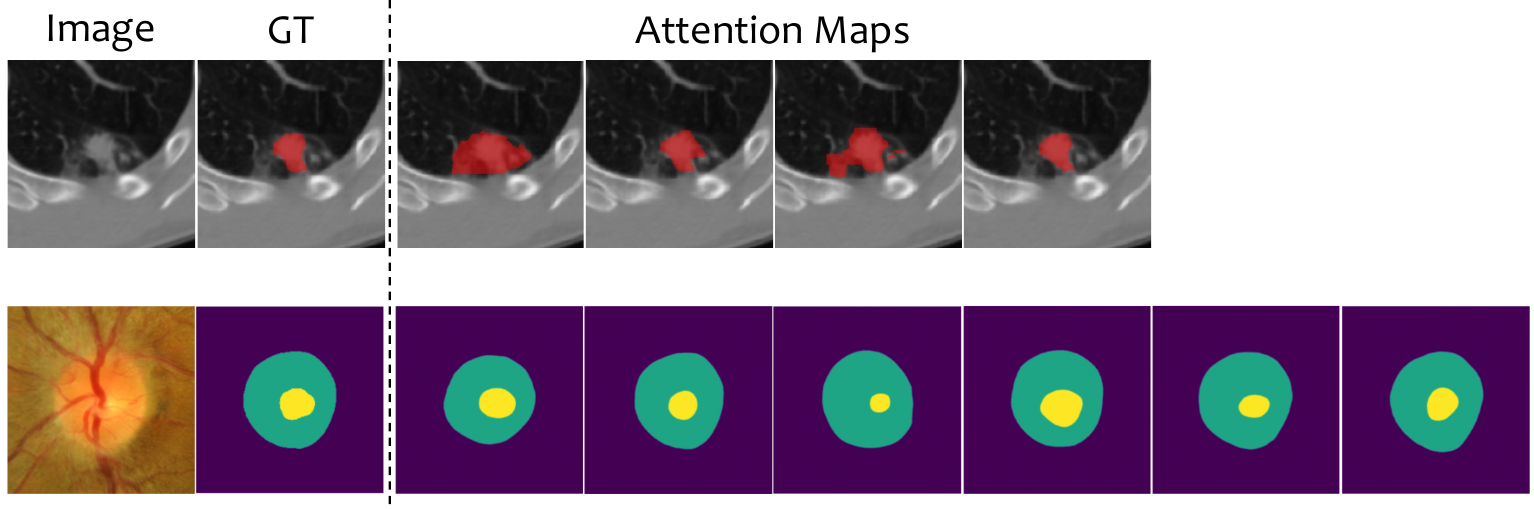}
\end{center}
\vspace{-0.8em}
\caption{Visualizations on groundtruth (GT) provided by senior doctors and noisy segmentation labels. Examples from LIDC-IDRI and RIGA dasets are shown respectively.} 
\label{fig:real_label}
\vspace{-1em}
\end{figure}

Following \cite{Zhang2020disentangling,Ji2021learning}, we evaluate LF-Net on datasets from varied image modalities: MNIST, ISBI2015 MS lesion \cite{Jesson2015hierarchical}, Lung node segmentation dataset (LIDC-IDRI) \cite{SG2011the}, RIGA \cite{Almazroa2017agreement}, and one hyperspectral image segmentation benchmark choledoch dataset \cite{Zhang2019a}. Many methods are evaluated with the help of synthetic images or labels \cite{Yang2015an,Han2018co,Wu2020accurate,Xu2022anti,Xue2022robust,Ju2022improving}. Strictly following the simulated annotation process in \cite{Zhang2020disentangling}, we simulate a range of annotator types on MNIST, MS lesion, and the choledoch datasets. LIDC-IDRI and RIGA datasets contain real experts' annotations for every image. We directly adopt these annotations as multi-rater annotations. Since the groundtruth masks are unknown in these two datasets, we ask one senior eye-doctor and one senior radiologist from a top-tier hospital to provide the groundtruth mask. The groundtruth mask and the noisy annotations are shown in Fig.~\ref{fig:real_label}.

\begin{figure}[t]
\begin{center}
    \includegraphics[width=1\linewidth]{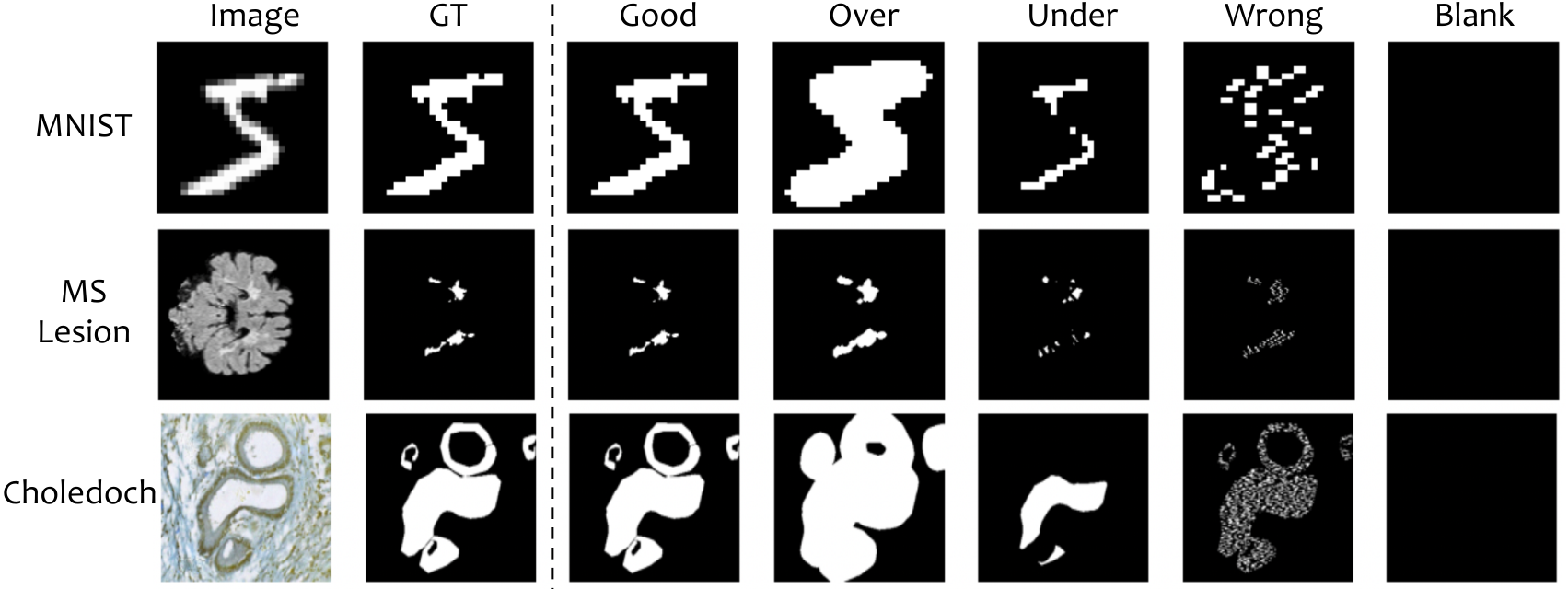}
\end{center}
\vspace{-0.8em}
\caption{Visualizations on five noisy segmentation labels. Original image and groundtruth are also shown on the left side of the dotted line as reference.} 
\label{fig:noisy_label}
\vspace{-1em}
\end{figure}

More concretely, following \cite{Zhang2020disentangling}, for MNIST dataset, we acquire the groundtruth segmentation labels by thresholding the pixel values at 0.5, \emph{i.e.}, pixels with values larger or equal than 0.5 will be assigned with class 1 and others with class 0. For MS lesion dataset \cite{Jesson2015hierarchical}, following \cite{Zhang2020disentangling}, we hold 10 scans for training and 11 scans for testing. We use 20\% of training images as a validation set for MNIST and MS lesion. Choledoch dataset \cite{Zhang2019a} consists of 538 images with 60 spectral bands. We randomly hold 438 images for training and the remaining 100 images for testing. Strictly follow \cite{Zhang2020disentangling}, we apply Morpho-MNIST software \cite{Castro2019quantitative} to generate five noisy segmentation labels: good, over, under, wrong and blank segmentation. Fig.~\ref{fig:noisy_label} shows some noisy label examples. 

In LIDC-IDRI dataset \cite{SG2011the}, there are four annotators per image from different experts. Noted that this dataset does not have consistent rater IDs across images. Following the pre-processing method in \cite{Kohl2018a}, we first split LIDC dataset into 722 patients' data for training, 144 for validation, and 144 for testing. Then we resample the CT scans to 0.5 $\times$ 0.5 mm$^2$ voxel spatial resolution. Lastly, we crop 2D images (180 $\times$ 180 pixels) centered at the lesion positions where at least one expert segmented a lesion so as to focus on the annotated lesions. RIGA dataset \cite{Almazroa2017agreement} is analyzed extensively in \cite{Ji2021learning} for retinal cup and disc segmentation with six label maps annotated by six glaucoma experts. Following \cite{Ji2021learning}, we hold 195 images from BinRushed and 460 images from MESSIDOR for training, and the remaining 95 images from Magrabia are used for testing. 

\subsection{Evaluation Metric}
The segmentation accuracy for MNIST, MS lesion, choledoch, and LIDC-IDRI datasets is measured by the well-known \underline{D}ice-\underline{S}{\o}rensen \underline{C}oefficient (DSC), and \underline{b}alanced \underline{A}verage \underline{H}ausdorff \underline{D}istance (bAHD) \cite{Aydin2020on}. For RIGA dataset, we follow the same evaluation metric in \cite{Ji2021learning}, \emph{i.e.}, soft DSC / \underline{I}ntersection \underline{o}ver \underline{U}nion (IoU) metrics through multiple threshold levels, set as (0.1, 0.3, 0.5, 0.7, 0.9), instead of using a single threshold (\emph{i.e.}, 0.5). Noted that in the setting of soft metrics, we assume there exist multiple underlying groundtruth maps. This does not contradict with our hypothesis that there exists only one groundtruth map, as the degree of annotation error is very low in RIGA dataset. All models are run for at least 3 times with different random initializations to compute the mean and standard deviation, as suggested in \cite{Zhang2020disentangling}.
 
\subsection{Implementation Details}
Our method is implemented in Pytorch. All networks used in LF-Net and competitors are 2D U-Net \cite{Ronneberger2015unet}, consisting of four encoder and decoder blocks, with the channel number of 32, 64, 128, 256, respectively, unless otherwise specified. In rater characteristics net, the backbone network is the 2D U-Net, except the last decoder layer, where the branches are split to modeling characteristics of the individual rater. For 3D data such as hyperspectral images or CT images, our LF-Net can also be built on 3D backbones, but this is out of the scope of this paper. Adam \cite{Kingma2015adam} optimiser is used in all experiments. Other network training hyper-parameters (and the default values we use) are batchsize (2), learning rate (1$\times$10$^{-4}$), $\lambda$ (0.05), $\tau$ (2.5), training epoch for the soft label learning network and the proposed segmentation network (10, 70). Since the image size in MNIST is small, different from other datasets, the batch size for MNIST is set to 128. $\lambda=0.01$ and the training epoch for 
rater characteristics net module is 15. All of the models are implemented and tested on a workstation with GeForce GTX 1080 GPU. We do not conduct any data augmentation for all datasets during training.

\subsection{Comparison between LF-Net and State-of-the-arts}
We conduct comparisons between LF-Net and eight competitors: an U-Net model trained with a single segmentation label obtained by taking the majority vote, the mean, and the STAPLE of all noisy labels are denoted as 1) MV-UNet, 2) Mean-UNet, and 3) STAPLE-UNet, respectively; 4) LS-UNet \cite{Jensen2019improving}; 5) MH-UNet \cite{Guan2018who}; 6) CM-UNet \cite{Zhang2020disentangling}; 7) MR-Net \cite{Ji2021learning}. We also compare with the proposed qualified majority vote (QMV) method, which directly adopt the vanilla segmentation network with the threshold voting scheme. The training epoch for MV-UNet, QMV-UNet, Mean-UNet and STAPLE-UNet is 150. We apply the default training strategies for other competitors. Noted that the main purpose of MR-Net \cite{Ji2021learning} is slightly different with ours. In order to show our effectiveness even compared with MR-Net, we conduct experiments on the RIGA dataset \cite{Almazroa2017agreement} used in MR-Net, and apply the same evaluation metric \emph{i.e.}, the soft DSC / IoU metrics introduced in MR-Net.

\begin{figure}[t]
\begin{center}
    \includegraphics[width=1\linewidth]{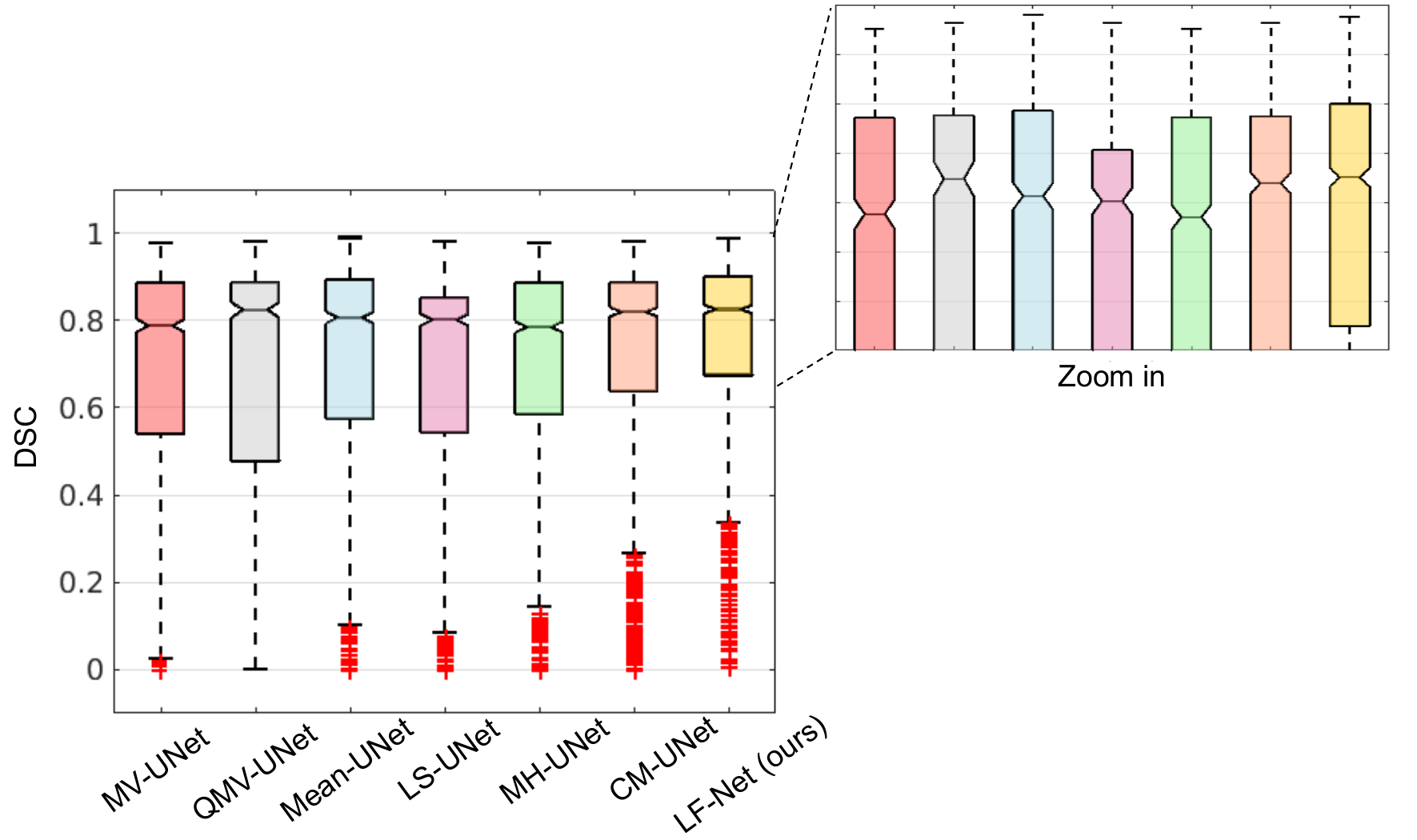}
\end{center}
\vspace{-0.7em}
\caption{DSC comparison in box plots on LIDC-IDRI dataset. Proposed LF-Net improves the overall mean and median DSCs. The $p$-values for testing significant difference between LF-Net and other methods are all less than 0.05.} 
\label{fig:boxplot}
\vspace{-1em}
\end{figure}

Quantitative results are summarized in Table~\ref{tab:comparison}. Since we cannot guarantee the annotations generated in our experiments are exactly the same as those used in \cite{Zhang2020disentangling}, we run the official codes released by \cite{Jensen2019improving,Guan2018who,Zhang2020disentangling} and obtain the results. To show the noisy level of simulated annotations in these datasets, \emph{Average} is also provided in Table~\ref{tab:comparison}, computed by taking the average DSC of all annotation maps with the groundtruth map. \emph{Oracle} serves as an upper bound, corresponding to training LF-Net by using the groundtruth annotation instead of noisy annotations. Table~\ref{tab:comparison} shows that our LF-Net outperforms all competitors by a large margin.
LF-Net achieves much better performance than these baseline models, indicating that directly train a deep network by the typical groundtruth label via \emph{e.g.}, majority vote may be over-confident \cite{Ji2021learning}, causing the over-fitting issue during training. LF-Net trained with multi-rater annotations is more robust towards raters' errors. Our soft label supervision provides more information to the segmentation network compared with only hard labels.

\begin{figure*}[t]
\begin{center}
    \includegraphics[width=0.87\linewidth]{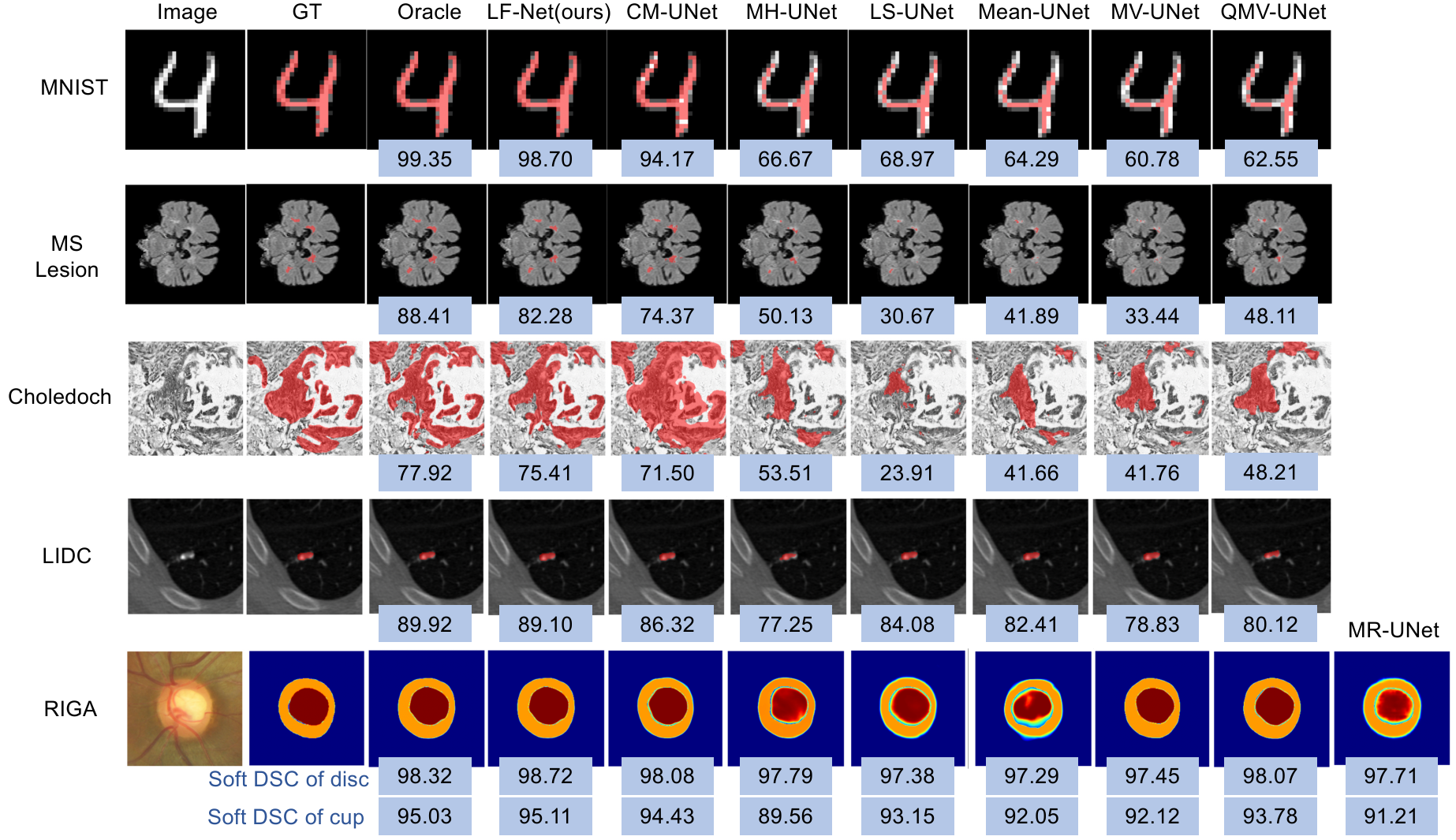}
\end{center}
\vspace{-0.7em}
\caption{(a) Qualitative comparisons on five datasets. Numbers on the bottom show segmentation DSCs. (b) DSC comparison in box plots on Choledoch dataset. Proposed LF-Net improves the overall mean and median DSCs. (c) Segmentation DSC (\%) of different models given different annotation noise levels (average DSC) on Choledoch dataset.} 
\label{fig:vis}
\vspace{-1em}
\end{figure*}

\begin{table}[!tb]
\renewcommand\arraystretch{1}
\footnotesize
\centering
\caption{Performance comparison (in DSC, $\%$ and bAHD) on LIDC-IDRI \cite{SG2011the} datasets. QMV-UNet indicates using our proposed qualified majority voting scheme. Experiments are repeated at least 3 times with different model initializations to compute the mean and standard deviation, as suggested in \cite{Zhang2020disentangling}.}
\label{tab:lidc}
\begin{tabular}{lcc}
\toprule[0.15em]
{Methods} & DSC~$\uparrow$ & bAHD~$\downarrow$\\
\midrule[0.09em]
MV-UNet & 67.11 $\pm$ 0.41 & 51.02 $\pm$ 3.31\\
QMV-UNet & 66.48 $\pm$ 0.87 & 43.94 $\pm$ 2.85 \\
Mean-UNet & 64.96 $\pm$ 0.69 & 53.52 $\pm$ 2.42 \\
LS-UNet \cite{Jensen2019improving} & 65.42 $\pm$ 1.61 & 47.62 $\pm$ 3.58 \\
MH-UNet \cite{Guan2018who} & 66.85 $\pm$ 1.01 & 45.46 $\pm$ 1.97\\
CM-UNet \cite{Zhang2020disentangling} & 68.63 $\pm$ 1.39 & 39.08 $\pm$ 2.54 \\
LF-Net (ours) & \textbf{73.35} $\pm$ \textbf{0.55} & \textbf{23.78} $\pm$ \textbf{2.82} \\
\arrayrulecolor{black!30}\midrule
\emph{Oracle} & 74.21 $\pm$ 0.52 & 21.93 $\pm$ 2.37 \\
\bottomrule[0.15em]
\end{tabular}
\end{table}

\begin{table}[!tb]
\renewcommand\arraystretch{1}
\footnotesize
\centering
\caption{Performance comparison (in soft DSC, $\%$ and soft IoU $\%$) on RIGA dataset \cite{Almazroa2017agreement}. QMV-UNet indicates using our proposed qualified majority voting scheme. Experiments are repeated at least 3 times with different model initializations to compute the mean and standard deviation, as suggested in \cite{Zhang2020disentangling}. ``$\uparrow$" indicate the larger the better. \textbf{Bold} denotes the best results per measurement. \emph{Oracle} means our LF-Net is trained with all groundtruth annotations, served as the upper bound.}
\label{tab:riga}
\resizebox{1\linewidth}{!}{
\begin{tabular}{lcccc}
\toprule[0.15em]
{Methods} & DSC$_\text{disc}^\text{soft} \uparrow$ & IoU$_\text{disc}^\text{soft} \uparrow$ & DSC$_\text{cup}^\text{soft} \uparrow$ & IoU$_\text{cup}^\text{soft} \uparrow$\\ 
\midrule[0.09em]
MV-UNet & 96.05 $\pm$ 0.11 & 92.37 $\pm$ 0.05 & 84.68 $\pm$ 0.05 & 73.43 $\pm$ 0.02\\
QMV-UNet & 97.23 $\pm$ 0.15 & 94.61 $\pm$ 0.24 & 85.12 $\pm$ 0.13 & 74.09 $\pm$ 0.21 \\
Mean-UNet & 96.14 $\pm$ 0.10 & 92.55 $\pm$ 0.20 & 81.67 $\pm$ 0.24 & 68.98 $\pm$ 0.38\\
LS-UNet \cite{Jensen2019improving} & 95.21 $\pm$ 0.03 & 90.87 $\pm$ 0.05 & 81.22 $\pm$ 0.28 & 68.41 $\pm$ 0.34\\
MH-UNet \cite{Guan2018who} & 96.23 $\pm$ 0.10 & 92.71 $\pm$ 0.18 & 81.47 $\pm$ 0.11 & 68.74 $\pm$ 0.17\\
CM-UNet \cite{Zhang2020disentangling} & 97.53 $\pm$ 0.06 & 95.17 $\pm$ 0.13 & 84.75 $\pm$ 0.26 & 74.80 $\pm$ 0.28\\
MR-UNet \cite{Ji2021learning} & 96.44 $\pm$ 0.21 & 93.10 $\pm$ 0.42 & 86.23 $\pm$ 0.11 & 75.78 $\pm$ 0.23 \\
LF-Net (ours) & \textbf{97.87} $\pm$ \textbf{0.06} & \textbf{95.75} $\pm$ \textbf{0.10} & \textbf{88.77} $\pm$ \textbf{0.11} & \textbf{79.70} $\pm$ \textbf{0.15} \\
\arrayrulecolor{black!30}\midrule
\emph{Oracle} & 97.73 $\pm$ 0.04 & 95.58 $\pm$ 0.07 & 88.47 $\pm$ 0.04 & 79.32 $\pm$ 0.08 \\
\bottomrule[0.15em]
\end{tabular}
}
\vspace{-0.6em}
\end{table}

Results for LIDC-IDRI \cite{SG2011the} dataset are reported in Table~\ref{tab:lidc}. It can be seen that the proposed LF-Net outperforms all competitors by a large margin, \emph{e.g.}, around 5$\%$ in DSC compared with CM-UNet, and 7$\%$ compared with MH-UNet. Noted that LIDC-IDRI dataset does not have consistent rater IDs across images, and many annotations are blank. We conduct an experiment on LF-Net without rater characteristics label supervision (\emph{i.e.}, \textit{w/o} rater characteristics net module), which achieves 73.27$\%$ in DSC (73.35$\%$ for LF-Net). We can see that for such a dataset, rater characteristics label supervision cannot help improve the segmentation performance. 

Fig.~\ref{fig:boxplot} shows comparison results of our LF-Net with other methods on LIDC-IDRI dataset by box plots. Besides, the $p$-values for testing significant difference between LF-Net and other methods are all less than 0.05, showing our LF-Net is significantly better than other competitors.

\begin{table}[t]
\renewcommand\arraystretch{1}
\footnotesize
\centering
\caption{Ablation on Choledoch \cite{Zhang2019a} and RIGA \cite{Almazroa2017agreement} dataset. {\color{red}${\dagger}$} indicates LF-Net without threshold voting scheme when trained with soft label supervision (no $m_{i,j}^\text{tv}$ in Eq.~\ref{eq:sls}). {\color{red}${\ddagger}$} means segmentation network is trained without $\mathcal{L}_\text{ce}$. S means the vanilla segmentation model supervised by only majority voting. TV means the threshold voting scheme. SLS means the segmentation network is supervised by soft label supervision. RCLS means the segmentation network is trained with rater characteristics label supervision (\textit{w/} rater characteristics net module). No threshold voting refers to the methods using majority-voted labels of all pixels for training.}
\label{tab:ablation-hsi}
\resizebox{1\linewidth}{!}{
\begin{tabular}{cccc|ccc}
\toprule[0.15em]
& & & & Choledoch & \multicolumn{2}{c}{RIGA}\\
\cmidrule(lr){5-5} \cmidrule(lr){6-7}
S & TV & SLS & RCLS & DSC (\%) & DSC$_\text{disc}^\text{soft}$ (\%) & DSC$_\text{cup}^\text{soft}$ (\%)\\
\midrule
\checkmark & & &  & 36.87 $\pm$ 0.73 & 96.08 $\pm$ 0.15 & 84.73 $\pm$ 0.07 \\ \rowcolor{gray!20}
\checkmark & \checkmark & & & 50.07 $\pm$ 0.52 & 97.23 $\pm$ 0.15 & 85.12 $\pm$ 0.13 \\
\checkmark & & \checkmark &  & 40.12 $\pm$ 1.22 & 96.20 $\pm$ 0.14 & 84.52 $\pm$ 0.29 \\\rowcolor{gray!20}
\checkmark & & & \checkmark & 54.02 $\pm$ 0.82 & 97.31 $\pm$ 0.11  & 87.37 $\pm$ 0.72\\
\checkmark & \checkmark & \checkmark &  & 63.27 $\pm$ 1.01 & 97.62 $\pm$ 0.06 & 88.16 $\pm$ 0.19\\\rowcolor{gray!20}
\checkmark & \checkmark & \checkmark & \checkmark & \textbf{65.35} $\pm$ 0.33 & \textbf{97.88} $\pm$ 0.06 & \textbf{88.77} $\pm$ 0.15\\
\checkmark & \checkmark{\color{red}${\dagger}$} & \checkmark & \checkmark & 64.72 $\pm$ 0.29 & 97.49 $\pm$ 0.07 & 88.71 $\pm$ 0.13\\\rowcolor{gray!20}
\checkmark{\color{red}${\ddagger}$} & \checkmark & \checkmark & \checkmark & 64.92 $\pm$ 0.18 & 97.64 $\pm$ 0.05 & 88.65 $\pm$ 0.16\\
\bottomrule[0.15em]
\end{tabular}
}
\vspace{-0.6em}
\end{table}

We also summarize the results on RIGA dataset in Table~\ref{tab:riga}. The final soft DSC and IoU are obtained by averaging the results of multiple thresholds, denoted as Dice$^\text{soft}$ and IoU$^\text{soft}$, the higher the better. Our method achieves superior performance compared with other methods, even the state-of-the-art method for joint optics and disc segmentation: MR-Net \cite{Ji2021learning}. For training MR-Net, we adopt its officially released codes, who utilizes the U-Net architecture with ResNet34 as the backbone. Noted that the U-Net architecture with ResNet34 is a much stronger backbone than ours. Although groundtruth labels are not utilized to train LF-Net, its performance outperforms \emph{oracle}, which use groundtruth labels during training. The reason may be that we use soft DSC / IoU metrics to do the evaluation, characterizing the agreement when threshold is high and the disagreement when threshold is low. In other words, soft metrics characterizes both agreement and disagreement among annotations. Our qualified majority voting strategy supervises the most confident pixels (agreement) and propagates knowledge to other pixels, while the rater characteristics net module models the disagreement among raters. Thus, compared with LF-Net, \emph{oracle} yields confident predictions, bringing negative influences on soft DSC /IoU. We show qualitative comparison in Fig.~\ref{fig:vis}. Compared with other methods, LF-Net ensures to learn the true label, which is more robust to the complicated background.

\begin{figure*}[t]
\begin{center}
    \includegraphics[width=0.9\linewidth]{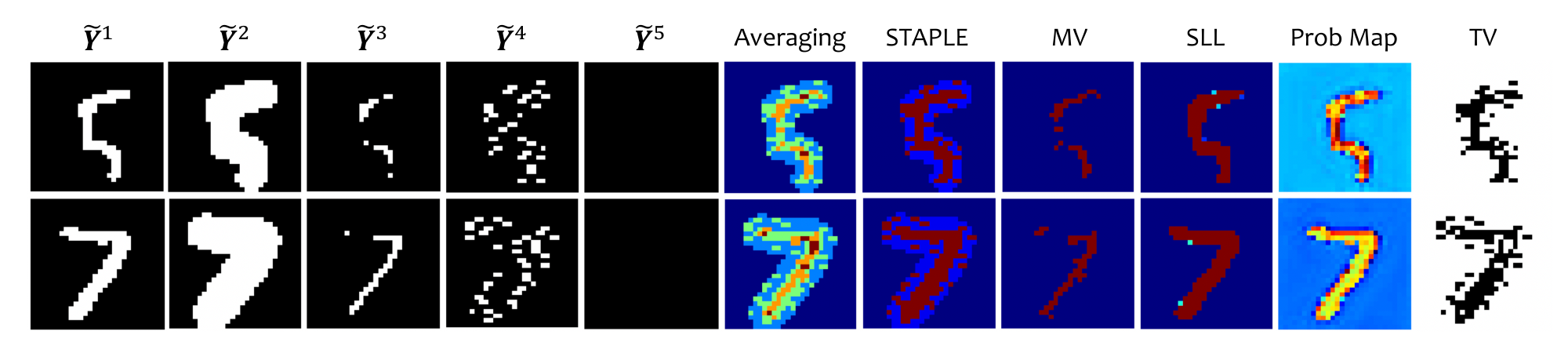}
\end{center}
\vspace{-0.7em}
\caption{Visualization of different fused labels on MNIST dataset. Binary masks are shown in binary maps, and soft labels are shown in heat maps. We randomly select two images and visualize their noisy annotations $\hat{\mathbf{Y}}^1$ - $\hat{\mathbf{Y}}^5$, where we drop footnote $i$ for simplicity. Averaging, STAPLE, MV and SLL denote the fused label obtained by averaging, STAPLE, majority vote and the proposed soft label learning. Prob Map means the probability map obtained by the segmentation network. TV is the threshold voting map $\mathbf{M}^{tv}$, where black denotes ``0'' and white denotes ``1''.} 
\label{fig:soft_label_learning}
\vspace{-0.5em}
\end{figure*}

\begin{figure}[t]
\begin{center}
    \includegraphics[width=0.59\linewidth]{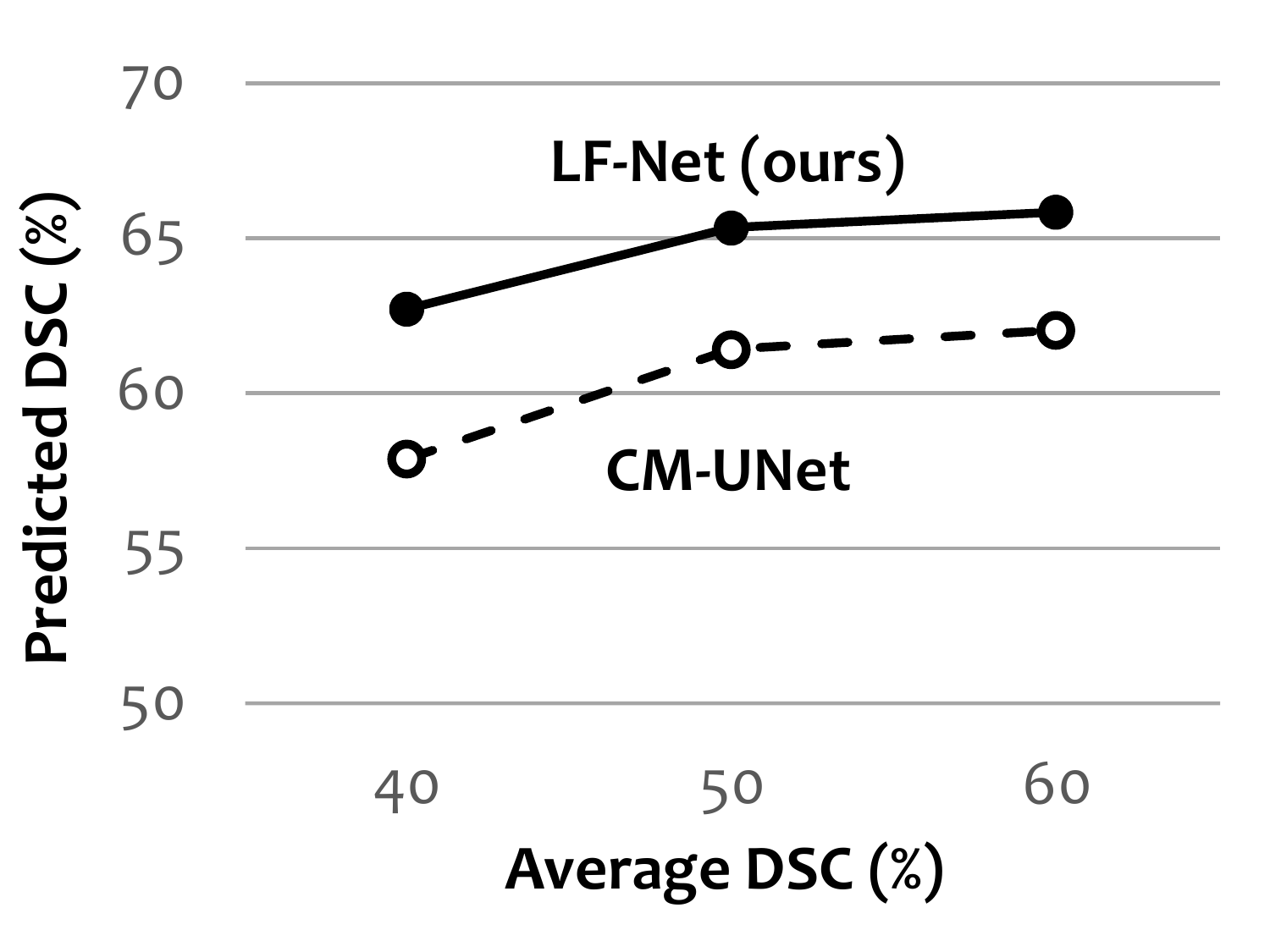}
\end{center}
\vspace{-0.7em}
\caption{Segmentation DSC (\%) of different models given different annotation noise levels (average DSC) on Choledoch dataset.} 
\label{fig:noise_level}
\vspace{-0.7em}
\end{figure}

\subsection{Ablation Study}
To demonstrate the indispensability of each module, we conduct ablation study on Choledoch dataset \cite{Zhang2019a} (with synthetic annotations) and RIGA dataset \cite{Almazroa2017agreement} (with real experts' annotations). As shown in Table~\ref{tab:ablation-hsi}, vanilla segmentation network with threshold voting scheme (see S + TV in the table) outperforms vanilla segmentation network \textit{w/} majority voting but \textit{w/o} threshold voting, indicating the training process supervised by the qualified majority voting (\emph{i.e.}, trustworthy labels of a subset of pixels) leads to performance gains. The segmentation network is trained with both qualitative majority voting strategy and soft label supervision (S + TV + SLS) yields significant improvements compared with the above models, showing the importance of soft supervision. LF-Net (S + TV + SLS + RCLS) further boosts the performance by modeling the characteristics of individual raters. Since LF-Net is trained with mixed supervision (see Eq.~\ref{eq:seg_loss}), if we remove $\mathcal{L}_\text{ce}$ while keeping the other auxiliary supervision, the results (the last row in Table~\ref{tab:ablation-hsi}) are slightly worse than LF-Net. 

It is worth mentioning that instead of estimating the pixel-wise confusion matrices on a per image basis as in \cite{Zhang2020disentangling}, our rater characteristics net module captures raters' characteristics in a more effective way. If replacing rater characteristics net module in LF-Net by the confusion matrix proposed by \cite{Zhang2020disentangling}, Choledoch dataset yields $64.58 \pm 0.63$ in DSC.

To show the fused soft label acquired from our soft label learning network is able to provide the segmentation network with more informative knowledge and is more helpful for training than other typical fused label, we replace soft label learning network by majority vote and STAPLE, obtaining only 54.17 $\pm$ 0.22 and 62.65 $\pm$ 1.36 in DSC (\%) on Choledoch dataset. To better understand why the proposed soft label learning network can provide better supervision, we also visualize the fused label obtained by averaging, STAPLE, majority vote, the proposed soft label learning in Fig.~\ref{fig:soft_label_learning} in MNIST dataset. It is interesting to observe that averaging obtains an ambiguous soft label (see Averaging in Fig.~\ref{fig:soft_label_learning}). STAPLE achieves better fused label than averaging (see STAPLE). But, since it only predicts fused label in a per-pixel manner, which inevitably ignores information from the context, leading to the occurrence of many discrete areas. Majority vote has critical criteria for the target pixels, which results in under-estimated fused labels (see MV). Thanks to the context modeling ability of a CNN, our proposed soft label learning network can obtain confident and more accurate fused labels (see SLL).

Additionally, we compare LF-Net with CM-UNet \cite{Zhang2020disentangling} on Choledoch dataset for different \textit{average} DSCs where labels are generated by a group of five simulated annotators. Average DSC indicates the noise level of annotations, the higher the less noisy. Fig.~\ref{fig:noise_level} shows LF-Net consistently surpasses the state-of-the-art CM-UNet. 

\subsection{Parameter Analysis}

\begin{figure}[t]
\begin{center}
    \includegraphics[width=1\linewidth]{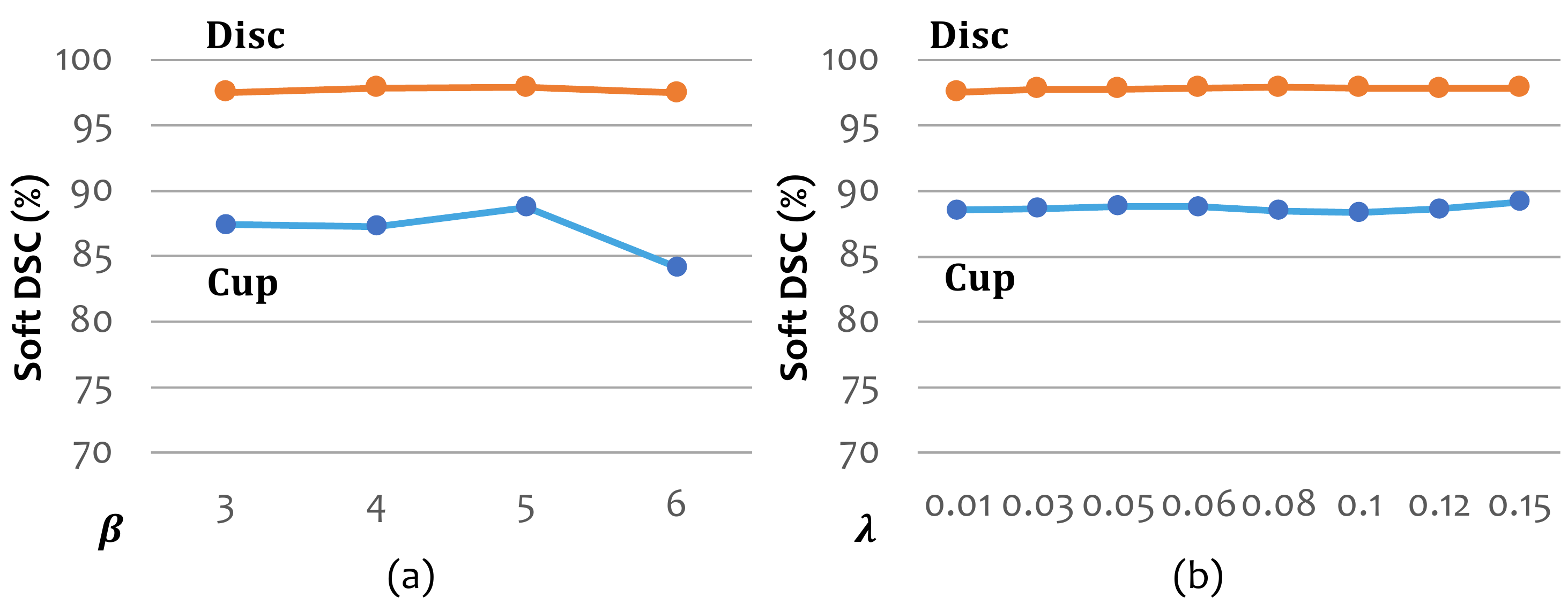}
\end{center}
\caption{Performance changes by varying (a) the threshold for threshold vote scheme, and (b) trade-off parameter for training the student model.} 
\label{fig:parameter}
\end{figure}

Two parameters are used in LF-Net, \emph{i.e.}, $\beta$ (Eq.~\ref{eq:tv}) and $\lambda$ (Sec.~\ref{sec:rcn}). We vary each of them and fix the other one to the default value to see how the performance changes on the RIGA dataset. We set $\beta=5$ and $\lambda = 0.05$ as default values. As shown in Fig.~\ref{fig:parameter}, the two parameters are not sensitive within certain ranges for RIGA dataset. When $\beta=R=6$, we can observe a performance drop for cup segmentation. This is because when $\beta=6$, the number of selected subset of pixels is too small to train a good model for cup.

\subsection{Discussion}
We are tackling a very important subject that is ubiquitous in medical imaging \cite{Xie2020recurrent,Dou2020unpaired,Li2018h,Tang2021weakly,Cai2021deep,Wang2021automatic,Song2017accurate,Wang2019blood}. In this work, we follow the scenario and the setting proposed in \cite{Zhang2020disentangling}. As suggested by \cite{Zhang2020disentangling}, MNIST, ISBI 2015 MS lesion and LIDC-IDRI datasets are adopted, where we strictly follow \cite{Zhang2020disentangling} to generate the noisy annotations in the first two datasets. Like many other multi-rater modeling works \cite{Kohl2018a,Kohl2019a,Ronneberger2015unet,Sohn2015learning,Jensen2019improving,Guan2018who,Yu2020difficulty}, our proposed approach requires most raters to segment each sample. 

Our LF-Net outperforms previous methods on synthetic datasets, which verifies that our LF-Net can better handle the situation when the provided annotations are with big noise levels (see Fig.~\ref{fig:noise_level}). For LIDC-IDRI, the average dice is 70.33$\%$. For RIGA dataset, the average DSC$_\text{disc}^\text{soft}$ is 98.33$\%$ and the average DSC$_\text{cup}^\text{soft}$ is 95.09$\%$. Compared with multi-rater agreement modeling work \cite{Ji2021learning} which mainly conducted experiments on real datasets, we synthesize noisy annotations in three datasets to show the effectiveness of our method when encountered with annotations with big noise levels. Compared with \cite{Zhang2020disentangling} which conducted experiments on datasets with synthetic annotations, we also test our method on datasets with real raters' annotations, showing the clinical practicality of the proposed LF-Net.

We term the annotation that provided by qualified majority voting as \emph{trustworthy} annotation. If there is no consensus achieved at the pixel, we simply disregard this pixel. In real scenarios, there will be pixels in some easy cases having qualified majority voting labels. It is out of the scope of this paper if no consensus achieved at any pixel in any case. Of course, these are extremes which rarely happen in practical. 

\section{Conclusion}
In this paper, we present \underline{L}abel \underline{F}illing framework (LF-Net) for medical image segmentation, given only noisy annotations during training. LF-Net selects a subset of pixels with trustworthy labels as supervision to train the network, guiding the direction of groundtruth label learning. To fill correct labels of other pixels while preventing over-fitting issues caused by the above process, two types of mix auxiliary supervision are proposed: a soft label learned by modeling the agreement knowledge from noisy annotations and rater characteristics labels which propagate individual rater's characteristics information. Experiment on five datasets from varied image modalities shows the superiority of LF-Net compared with state-of-the-art methods.

\bibliographystyle{IEEEtran}
\bibliography{bare_jrnl}

\end{document}


\title{Label Filling via Mixed Supervision for Medical Image Segmentation from Noisy Annotations}
\author{Ming Li, Wei Shen, Qingli Li, Yan Wang
\thanks{Manuscript submitted on July 2023. This work was supported in part by National Natural Science Foundation of China under Grant 62101191 and Grant 61975056, in part by Natural Science Foundation of Shanghai under Grant 21ZR1420800, and in part by the Science and Technology Commission of Shanghai Municipality under Grant 20440713100. }
\thanks{M. Li, Q. Li, and Y. Wang are with Shanghai Key Laboratory of Multidimensional Information Processing, East China Normal University, Shanghai, China (e-mail:10182100316@stu.ecnu.edu.cn, qlli@cs.ecnu.edu.cn, ywang@cee.ecnu.edu.cn). }
\thanks{W. Shen is with MoE Key Lab of Artificial Intelligence, AI Institute, Shanghai Jiao Tong University, Shanghai, China (e-mail: wei.shen@sjtu.edu.cn).}
\thanks{Corresponding author: Y. Wang}}

\markboth{Journal of \LaTeX\ Class Files,~Vol.~14, No.~8, June~2023}%
{Shell \MakeLowercase{\textit{et al.}}: A Sample Article Using IEEEtran.cls for IEEE Journals}

\newcommand{\yan}[1]{ \textcolor{red}{(yan: #1)}  }
\maketitle

{\appendices
\section{Proof for Lemma in Sec.~III-B}

In this section, we give \emph{proof} for the Lemma in Sec.~III-B.


\noindent \emph{Proof.}

Notice that for $i\in\{1,2,...,L\}$, we have
\begin{equation}
\begin{split}
    1-p_r = p(\tilde{y}_s^r\neq i|y_s=i)&=\sum_{j\neq i}p(\tilde{y}_s^r=j|y_s=i)\\
    &\geq p(\tilde{y}_s^r=t|y_s=i),
\end{split}\tag{A.1}
\end{equation}
for all $t\in\{1,2,...,L\}/\{i\}$.
Notice that 
\begin{equation}
    \begin{split}
        p(\tilde{y}_s^r=l)&=\sum_{i=1}^Lp(\tilde{y}_s^r=l|y_s=i)p(y_s=i)\\
        &=\frac{1}{L}\times(p_r+\sum_{i\neq l}p(\tilde{y}_s^r=l|y_s=i)).
    \end{split}\tag{A.2}
\end{equation}
Hence, we can obtain
\begin{equation}
\begin{split}
p(\tilde{y}_s^r=l)&\leq \frac{1}{L}\times(p_r+(L-1)\times(1-p_r))\\
&=\frac{L-1}{L}-\frac{L-2}{L}p_r.
\end{split}\tag{A.3}
\end{equation}
So,
\begin{equation}
\begin{split}
    p(\tilde{y}_s^r=l|y_s\neq l)&=\frac{p(y_s\neq l|\tilde{y}_s^r=l) p(\tilde{y}_s^r=l))}{p(y_s\neq l)}\\
    &=\frac{(1-p(y_s=l|\tilde{y}_s^r=l))p(\tilde{y}_s^r=l)}{p(y_s\neq l)}\\
    &=\frac{p(\tilde{y}_s^r=l)}{p(y_s\neq l)}-\frac{p(y_s=l|\tilde{y}_s^r=l)p(\tilde{y}_s^r=l)}{p(y_s\neq l)}\\
    &=\frac{p(\tilde{y}_s^r=l)}{p(y_s\neq l)}-\frac{p(\tilde{y}_s^r=l|y_s=l)p(y_s=l)}{p(y_s\neq l)}\\
    &\leq \frac{\frac{L-1}{L}-\frac{L-2}{L}p_r}{1-\frac{1}{L}}-\frac{p_r\times\frac{1}{L}}{1-\frac{1}{L}}\\
    &=1-p_r.
\end{split}\tag{A.4}
\end{equation}

\section{Proof for Theorem in Sec.~III-B}
In this section, we further give \emph{proof} for the Theorem in Sec.~III-B. 

\noindent \textbf{Theorem.} Assume for a pixel $s$, $p_{max}=\max\{p_1, p_2, ...,p_R\}<1$, $p_{min} = \min\{p_1,p_2,...,p_R\}>1/2$, and $\forall r=1,2,...,R$, $p(\tilde{y}_s^{r}\neq\tilde{l}|y_s=\tilde{l})>0$, where $p_r$ indicates the probability of labeling pixel $s$ correctly. $T$ raters label $s$ to class $\tilde{l}$. $T = R-C$, where $C$ is a constant, and $C\in\mathbb{N}$.  We define the probability that $y_s=\tilde{l}$ given R raters' annotations is ${P}_{gt}(R) = p(y_s=\tilde{l}|\tilde{y}_s^{x_1}=...=\tilde{y}_s^{x_T}=\tilde{l}, \tilde{y}_s^{x_{T+1}}, ..., \tilde{y}_s^{x_R}\in\{1,...,L\}/\{\tilde{l}\}, T=R-C)$. Then $\lim_{R\rightarrow +\infty}{P}_{gt}(R)=1$.

\noindent \emph{Proof.} 
\begin{equation}
\footnotesize
\begin{split}
    p(y_s|\tilde{y}_s^{x_1},...,&\tilde{y}_s^{x_R})=\frac{p(y_s)p(\tilde{y}_s^{x_1},...,\tilde{y}_s^{x_R}|y_s)}{p(\tilde{y}_s^{x_1},...\tilde{y}_s^{x_R}|y_s)p(y_s)+p(\tilde{y}_s^{x_1},...\tilde{y}_s^{x_R}|\bar{y}_{s})p(\bar{y}_{s})}\\
    &=\frac{p(y_s)\prod_{r=1}^{R}p(\tilde{y}_s^{x_r}|y_s)}{p(y_s)\prod_{r=1}^{R}p(\tilde{y}_s^{x_r}|y_s) + p(\bar{y}_{s})\prod_{r=1}^{R}p(\tilde{y}_s^{x_r}|\bar{y}_{s})}\\
\end{split}\tag{B.1}
\end{equation}
Then we can obtain the following lower bound:
\begin{equation}
\begin{split}
\footnotesize
{P}_{gt}&(R)=\frac{1}{1+(L-1)\frac{\prod_{r=1}^{T}p(\tilde{y}_s^{x_r}=\tilde{l}|y_s\neq\tilde{l})\prod_{r=T+1}^{R}p(\tilde{y}_s^{x_r}\neq\tilde{l}|y_s\neq\tilde{l})}{\prod_{r=1}^{T}p(\tilde{y}_s^{x_r}=\tilde{l}|y_s=\tilde{l})\prod_{r=T+1}^{R}p(\tilde{y}_s^{x_r}\neq\tilde{l}|y_s=\tilde{l})}}\\
&\geq \frac{1}{1+(L-1)\frac{\prod_{r=1}^{T}(1-p_{x_r})\prod_{r=T+1}^{R}p(\tilde{y}_s^{x_r}\neq\tilde{l}|y_s\neq\tilde{l})}{\prod_{r=1}^{T}p_{x_r}\prod_{r=T+1}^{R}p(\tilde{y}_s^{x_r}\neq\tilde{l}|y_s=\tilde{l})}}\\
&\geq \frac{1}{1+(L-1)\left(\frac{1-p_{min}}{p_{min}}\right)^{R-C}\frac{\prod_{r=R-C+1}^{R}p(\tilde{y}_s^{x_r}\neq\tilde{l}|y_s\neq\tilde{l})}{\prod_{r=R-C+1}^{R}p(\tilde{y}_s^{x_r}\neq\tilde{l}|y_s=\tilde{l})}},
\end{split}\tag{B.2}
\end{equation} 
where we use lemma in line 2. Since when $R\to + \infty$, $(\frac{1-p_{min}}{p_{min}})^{R-C} \to 0$ and
$\frac{\prod_{r=R-C+1}^{R}p(\tilde{y}_s^{x_r}\neq\tilde{l}|y_s\neq\tilde{l})}{\prod_{r=R-C+1}^{R}p(\tilde{y}_s^{x_r}\neq\tilde{l}|y_s=\tilde{l})}$ is bounded.
So we have $\lim_{R\rightarrow +\infty}\mathcal{P}_{gt}(R)\geq 1$. This completes the proof. $~~\qedsymbol$

\section{Derivation of the Probability for Sec.~III-B}
In this section, we give the derivation of the probability mentioned in Sec. III-B. When $L=2$, we assume that $l\in\{0, 1\}$. We now give a calculable lower bound of ${P}_{gt}(R)$. Note that for $r \in \{1,2,...,R\}$ we have
\begin{equation}
    p(\tilde{y}_s^r\neq l|y_s\neq l)=p(\tilde{y}_s^r= 1-l|y_s=1-l)=p_r,\tag{C.1}
\end{equation}
\begin{equation}
     p(\tilde{y}_s^r\neq l|y_s= l)=p(\tilde{y}_s^r= 1-l|y_s=l)=1-p_r.\tag{C.2}
\end{equation}
So we can obtain
\begin{equation}
\begin{split}
    {P}_{gt}(R)&\geq \frac{1}{1+(L-1)\left(\frac{1-p_{min}}{p_{min}}\right)^{R-C}\frac{\prod_{r=R-C+1}^{R}p(\tilde{y}_s^{x_r}\neq\tilde{l}|y_s\neq\tilde{l})}{\prod_{r=R-C+1}^{R}p(\tilde{y}_s^{x_r}\neq\tilde{l}|y_s=\tilde{l})}}\\
    &= \frac{1}{1+(\frac{1-p_{min}}{p_{min}})^{R-C}\frac{\prod_{r=R-C+1}^{R}p_{x_r}}{\prod_{r=R-C+1}^{R}(1-p_{x_r})}}\\
    &\geq \frac{1}{1+(\frac{1-p_{min}}{p_{min}})^{R-C}(\frac{p_{max}}{1-p_{max}})^C}.
\end{split} \tag{C.3}
\end{equation}
Thus, when $C=1$, $p_{max}=0.95$ and $p_{min}=0.75$, one can easily compute ${P}_{gt}(7)\geq 0.975$, ${P}_{gt}(8)\geq0.991$, and ${P}_{gt}(9)\geq0.997$.
}